\documentclass[runningheads]{llncs}

\usepackage{eccv}

\usepackage{eccvabbrv}

\usepackage{graphicx}
\usepackage{placeins}
\usepackage{booktabs}
\usepackage{url}
\usepackage{wrapfig}
\usepackage{graphicx}
\usepackage{enumitem}
\usepackage{mathtools}
\usepackage{array}
\usepackage{adjustbox}

\usepackage{booktabs}
\usepackage{multirow}
\usepackage{makecell}
\usepackage{amssymb}
\usepackage{appendix}
\usepackage{xcolor}
\usepackage{wrapfig}

\usepackage[accsupp]{axessibility}  % Improves PDF readability for those with disabilities.

\usepackage{hyperref}

\usepackage{orcidlink}
\usepackage{marvosym}

\begin{document}

\title{Beyond Inpainting: Unleash 3D Understanding for Stable Camera-Controlled Video Re-rendering}

\titlerunning{Beyond Inpainting}

\author{
Dong-Yu Chen \orcidlink{0009-0001-4301-3341}
, Yixin Guo \orcidlink{0009-0009-2554-469X}, Shuojin Yang\thanks{Corresponding author. } \orcidlink{0009-0009-5345-7972} \\ Tai-Jiang Mu \orcidlink{0000-0002-9197-346X} , Shi-Min Hu \orcidlink{0000-0001-7507-6542}
}
\authorrunning{Chen.~ et al.}

\institute{Key Laboratory of Pervasive Computing, Ministry of Education \\ Department of Computer Science and Technology, Tsinghua University, \\ Beijing 100084, China \\
 % \email{\{cdy24, guoyixin23\}@mails.tsinghua.edu.cn, \{yangshuojin, taijiang, shimin\}@tsinghua.edu.cn}
\email{yangshuojin@tsinghua.edu.cn}
} 

% \email{\{cdy24, guoyixin23\}@mails.tsinghua.edu.cn, \{yangshuojin, taijiang, shimin\}@tsinghua.edu.cn}

\maketitle

\newcommand{\teaser}{
% \centering
% \vspace{-25pt}
% {Project webpage:} \url{https://eleanor6725.github.io/DepthDirector/}
\includegraphics[width=\textwidth]{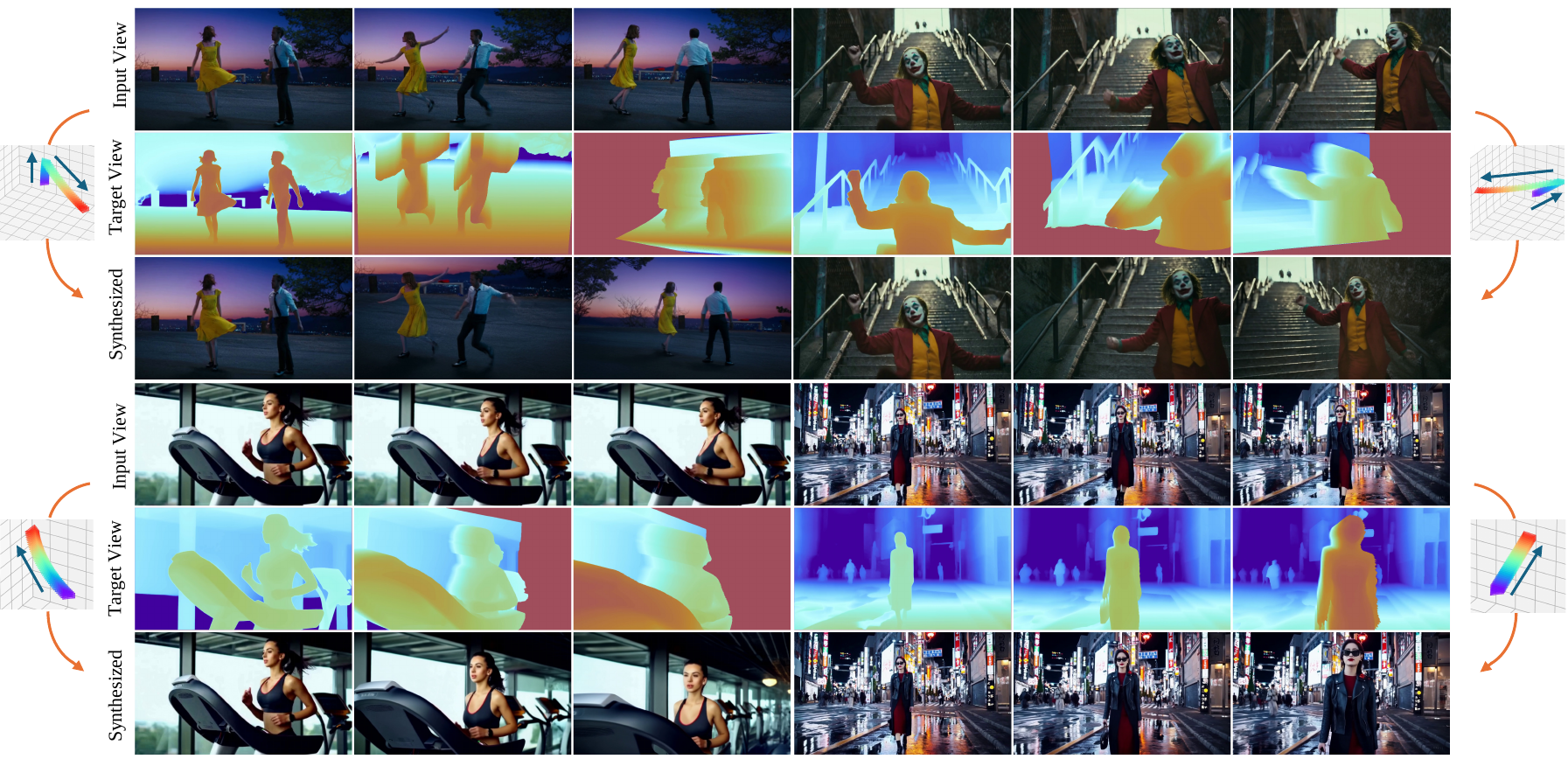}
% \vspace{-2em}
\captionof{figure}{\textbf{Example results synthesized by DepthDirector.} 
DepthDirector re-shoots the source video with novel camera trajectories. By fully leveraging the 3D understanding ability of video diffusion models, we are the first framework that achieves both precise camera controllability and consistent content preservation. % We visualized the novel camera trajectories alongside the video frames. 
}
% \vspace{-1em}
% \vspace{0.7em}
\label{fig_1}
}

\begin{abstract}

Camera control has been extensively studied in conditional video generation; however, stable altering camera trajectories while faithfully preserving video content remains a challenging task. 
The mainstream approach to achieving camera control is warping a 3D representation according to the target trajectory. 
However, such methods fail to fully leverage the 3D priors of video diffusion models (VDMs) and often fall into the Inpainting Trap, resulting in subject inconsistency and degraded generation quality. To address this problem, we propose DepthDirector, a video re-rendering framework with stable camera controllability. Our key insight is to decompose the conditional injection of video content and camera trajectories by leveraging the depth video from explicit 3D representation as camera-control guidance. 
Specifically, we design a View-Content Dual-Stream Condition mechanism, enabling VDMs to comprehend camera movements and leverage their 3D understanding capabilities. 
Additionally, we construct a large-scale multi-camera synchronized dataset named MultiCamWarp Dataset using Unreal Engine 5. Extensive experiments show that DepthDirector outperforms existing methods. Our code and dataset will be publicly available on our project website: \url{https://eleanor6725.github.io/DepthDirector/}.

\end{abstract}

\section{Introduction}
\label{sec:intro}

% meaning
Recent advances in video generation, particularly video diffusion models (VDMs), have enabled high-quality, content-controllable video synthesis from diverse inputs such as text, images, and videos \cite{animatediff,cogvideox,Wan}. 
Within this rapidly evolving field, camera control is crucial for generating expressive and cinematic videos. 
It not only helps content creators and cinematographers in framing scene content and dictating global motion, but also craft specific atmospheres and emphasize character emotions.
However, stable controllability of camera trajectories while faithfully preserving video content remains a significant challenge. 

% challenge % current methods
% implicit
Currently, the straight-forward solution to achieving camera control is to directly use camera parameters as implicit input conditions\cite{gcd, syncammaster, ReCamMaster, ac3d, cameractrl, cami2v}, encoding position and orientation through ray embeddings\cite{ac3d} or relative pose encoder\cite{ReCamMaster,syncammaster}. 
These control mechanisms attempt to tame VDMs with camera movements, but fail to bridge the gap between the user-desired camera trajectories and the implicit 3D geometry from video latents. This results in suboptimal physical consistency and camera instability. 
{Worse still, to learn the correspondence between camera parameters and scene geometry, such methods often rely on large-scale, multi-view synchronized rendered datasets \cite{syncammaster, gcd, ReCamMaster, camclonemaster} (e.g., \textbf{136K} videos for ReCamMaster) to implicitly learn 3D consistency. 
In contrast, we propose a new method that achieves superior performance with merely about $\textbf{5.8\%}$ of the training cost (using \textbf{8K} video sequences).}

\begin{figure}[!t]
    \centering
    \includegraphics[width=\textwidth]{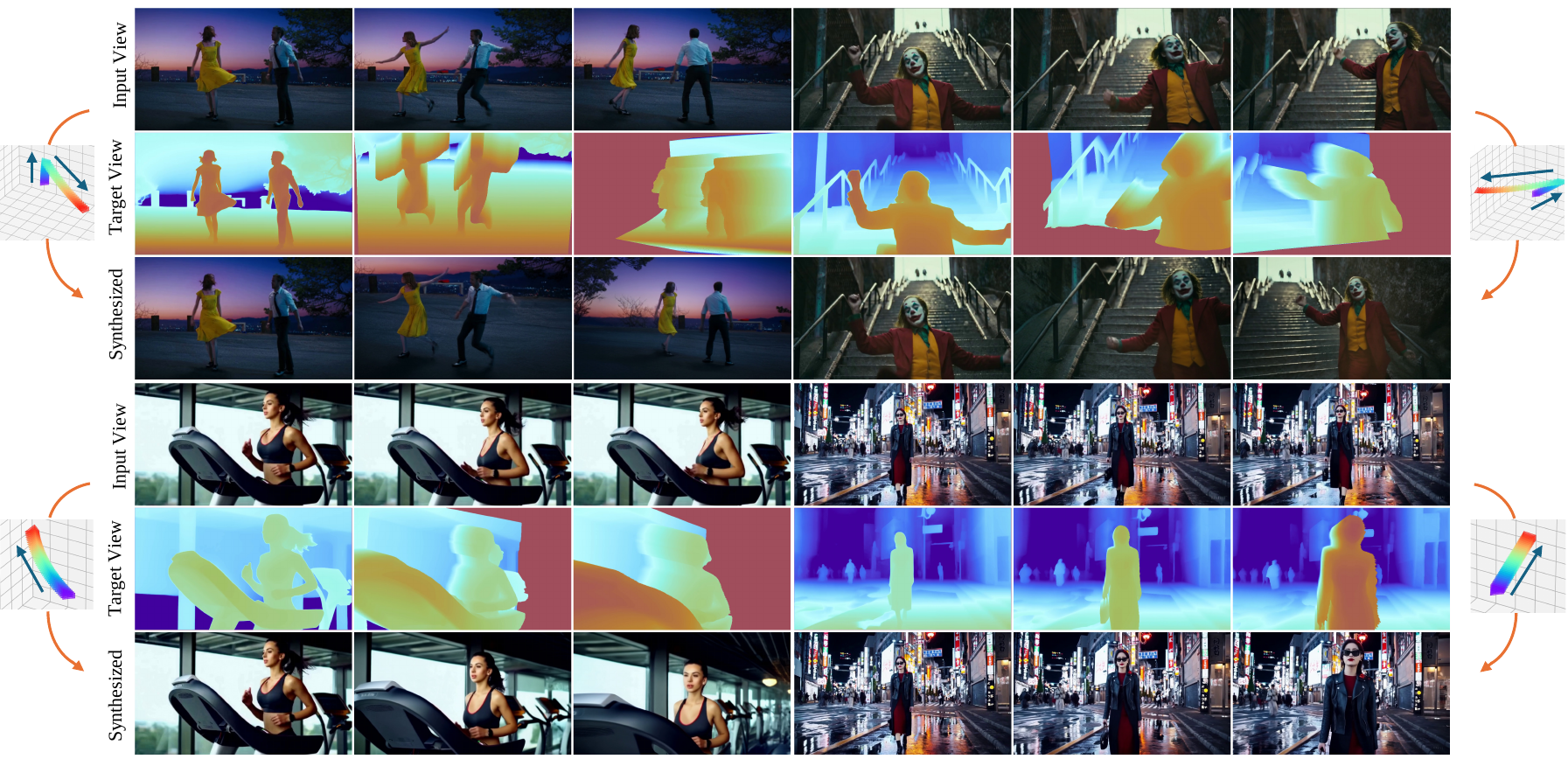}
    % \vspace{-2em}
    \captionof{figure}{\textbf{Example results synthesized by DepthDirector.} 
    DepthDirector re-shoots the source video with novel camera trajectories. 
    % By fully leveraging the 3D understanding ability of video diffusion models, we are the first framework that achieves both stable camera controllability and consistent content preservation. % We visualized the novel camera trajectories alongside the video frames. 
    }
    % \vspace{-1em}
    % \vspace{0.7em}
    \label{fig_1}
\end{figure}

\begin{figure}[!t]
    \centering
    \includegraphics[width=1.0\linewidth]{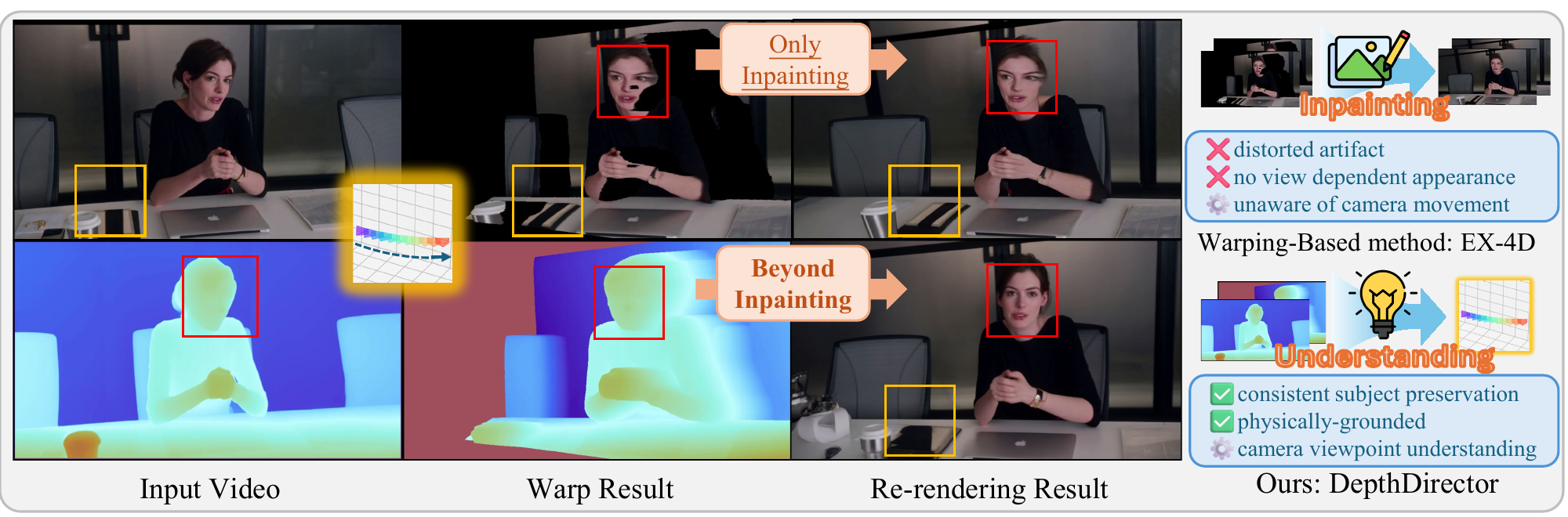}
    \caption{\textbf{Limitations of warping-based methods}. {Reprojected pixels exhibit noisy artifacts due to inaccurate 3D geometry. It leads to unrecoverable distortion to the identity and details of the subject, especially on human faces(See Red Box). Reflections on the table are baked as textures(Yellow Box). These \textbf{inpainting} artifacts are attributed to the unawareness of camera movements and video content. }}
    \label{fig:warp}
    % \vspace{-15pt}
\end{figure}

% explicit
Another line of methods \cite{ren2025gen3c,das,TrajectoryCrafter,ex4d,epic,worldforge} can achieve stable control of camera trajectories through explicit geometry-based guidance and generate final results via video \textbf{inpainting}.
However, this mechanism fundamentally caps the quality of the generated content at that of the warped RGB video. 
As illustrated in Fig.~\ref{fig:warp}, this leads to two primary issues. First, distortion: inevitable inaccuracies in 3D geometry estimated from monocular videos introduce distortions and artifacts into the warped RGB sequence, subsequently causing subject inconsistency and degraded fidelity.
Second, baked-in appearance: warped RGB sequence superficially “painting” subject appearance into textures, forcing the network to ignore the view dependent effects. 
We attribute these artifacts to the \textbf{Inpainting Trap},  a shortcut in VDMs that bypasses the understanding of the 3D/4D world and camera transformations. This reduces a VDM with 3D priors into a naive video inpainter.
To address this, we focus on designing a new geometry-based guidance mechanism that enables VDMs to internalize camera movements and faithfully preserve source content with physical consistency. 

To this end, we present DepthDirector, a framework that fundamentally redefines the camera-controlled video re-rendering paradigm, shifting from pixel-space inpainting to structure-guided synthesis. 
Our key insight is to \textbf{decompose} the conditional injection of video content and camera trajectories. This decomposition forces the network to implicitly fuse both of the conditions and resolve the novel-view extrapolation from its understanding of video content and camera movements. 
% From this perspective, warping-based methods fuse appearance information and novel camera viewpoint into a single warped RGB video, leading to an entangled guidance for VDMs to follow. 
To achieve this, we leverage temporally consistent warped depth rendered from explicit 3D representations under the target view as camera control guidance. This offers three key advantages: (1) It serves as the camera condition without appearance leakage. (2) 
It provides geometric information assisting VDMs in building a 3D/4D world understanding for input video. (3) It provides users with explicit and easy-to-operate camera controllability. 

More specifically, DepthDirector proposes a View-Content Dual-Stream Condition mechanism. A View Stream injects warped depth sequences into the denoised tokens, ensuring precise alignment with desired camera viewpoint. 
Simultaneously, a Content Stream utilizes the source video injection to faithfully preserve scene structure, semantics, dynamic motions and intrinsic properties. 
This decomposition ensures the integration of VDMs’ world-understanding prior to re-rendering, leading to stable camera controllability and consistent content preservation that goes \textbf{beyond inpainting}. 
As shown in Fig.~\ref{fig:warp}, by introducing the decomposed conditional mechanism, face identity is effectively preserved even in challenging human-centric scenes. In contrast, warping-based methods fuse appearance and camera viewpoint information into a single warped RGB video, leading VDMs to bypass understanding and solely inpaint from this entangled condition. This explains why methods\cite{TrajectoryCrafter} that adopt source video injection still fail to maintain consistent content preservation. 

% we introduce a lightweight LoRA-based video diffusion adapter to fine-tune our framework. This efficiently integrates geometric information from the warped-depth condition into the pretrained video diffusion model, producing visually coherent and realistic results while keeping computational costs manageable.

To train our new model, we construct a scalable multi-camera synchronized video dataset construction pipeline using Unreal Engine 5, named MultiCamWarp Dataset. This pipeline automatically renders realistic dynamic videos with ground truth depth in synchronized multi-view cameras. Using this system, we generate $\mathbf{8K}$ videos shot from $\mathbf{1K}$ different dynamic scenes. This dataset enables our model to effectively learn the multi-view consistency and camera controllability with much less training cost.
Experimental results show that our method outperforms state-of-the-art approaches and achieves both stable camera controllability and consistent content preservation. 

% TWO INSIGHTS
% a condition signal
% decomposition of camera information and content
%Our contributions can be summarized as follows:
%\begin{itemize}
%\item We propose a dual-stream injection mechanism that feeds both the source video and the warped depth sequence under the target camera into the video diffusion model, enabling the model to simultaneously comprehend scene geometry and dynamic content, thereby enhancing camera controllability and video content consistency.

%\item We propose Depth Director, a camera-controlled generative video re-rendering framework achieving both precise camera control and consistent subject preservation. Our experiments show Depth Director outperforms existing methods. 

%\item To enable the model to better learn the relationship between camera control, depth perception, and dynamic content, we develop a multi-camera synchronized video dataset using Unreal Engine 5, which contains $\mathbf{8K}$ realistic videos with ground truth depth, shoot from $\mathbf{1K}$ different dynamic scenes. 
     
%     % We will opensource the rendering code.
%\end{itemize}
\section{Related Works}
\label{sec:related}

\noindent\textbf{Video Diffusion Models.}
Video generation model has evolved from early approaches like Make-A-Video~\cite{Make-A-Video} and Gen-1~\cite{Gen-1} to more sophisticated models. SVD~\cite{svd} and VideoCrafter~\cite{videocrafter1,videocrafter2} enhanced temporal coherence, while large-scale models such as Hunyuan Video~\cite{hunyuanvideo}, CogVideoX~\cite{cogvideox}, and Wan ~\cite{Wan} achieve impressive spatiotemporal consistency and 3D understanding ability. 

\noindent\textbf{Video Depth Estimation}
Recovering depth from monocular video has been widely studied. A line of works focuses on estimating temporally consistent depth without camera estimation. VDA\cite{VDA} builds upon transformer model and enforce temporal consistency by additional loss. DepthCrafter\cite{DepthCrafter} and ChronoDepth\cite{chronodepth} finetune video diffusion models\cite{svd} to yield high-quality depth sequences. GeometryCrafter\cite{geocrafter} extends this paradigm to predict per-frame point map enabling camera intrinsic estimation. Another line of works\cite{wang2025pi3, wang2025vggt, map-anything, dens3r, movies, xu2026towards} directly regresses per-frame depth maps and camera parameters in a feedforward manner. These methods leverage multi-task learning and large-scale datasets to achieve superior performance; however, the estimated depth maps still suffer from low resolution and inaccuracy.

\noindent\textbf{Implicit Camera Conditioned Video Generation}
In camera-controlled video generation \cite{direct_a_video, vd3d, cami2v, zheng2025vidcraft3, animatediff}, researchers aim to directly incorporate camera parameters into video generation models to control the output viewpoint video. Camera extrinsics are injected into the diffusion model via direct parameter concatenation~\cite{motionctrl}, Pl\"ucker Embedding~\cite{lfns, cameractrl, CVD, ac3d, vd3d}, training camera encoder~\cite{gcd, syncammaster, ReCamMaster} and reference video~\cite{camclonemaster}. Since the correspondence between camera parameters and generated views isn't straight-forward, they ~\cite{gcd, camclonemaster, ReCamMaster, syncammaster} typically rely on large-scale multi-view rendered datasets \cite{syncammaster, gcd, ReCamMaster, camclonemaster} to implicitly learn 3D consistency and suffer from substantial camera control instabilities. 

\noindent\textbf{Explicit Camera Conditioned Video Generation}
To leverage explicit geometry reconstructed from input videos and incorporate precise camera conditioning, existing methods achieve camera control by using an anchor video~\cite{das,ren2025gen3c,TrajectoryCrafter,epic,flovd,recapture,voyager}. Some earlier works~\cite{gs-dit, trajattn, das} leverage 3D point tracking~\cite{spatialtracker,cotracker} to inject camera movements, which can cause artifacts if tracking fails. In order to inject 3D information, another line of works~\cite{TrajectoryCrafter,ren2025gen3c,ex4d,hou2024training,epic,recapture} uses a warp-and-inpaint strategy. They reconstruct a per-frame 3D representation and repaint on a warped video of target trajectories. However, these methods are prone to produce artifacts from noisy warps. A concurrent work WorldForge~\cite{worldforge} designs a training-free method to mitigate the noisy artifacts, while our method introduces a new conditioning mechanism that injects camera viewpoints informed by explicit geometry, enabling higher video generation quality and achieving stable camera control.

\section{Method}

\begin{figure*}[!t]
    \centering
    \includegraphics[width=1.0\linewidth]{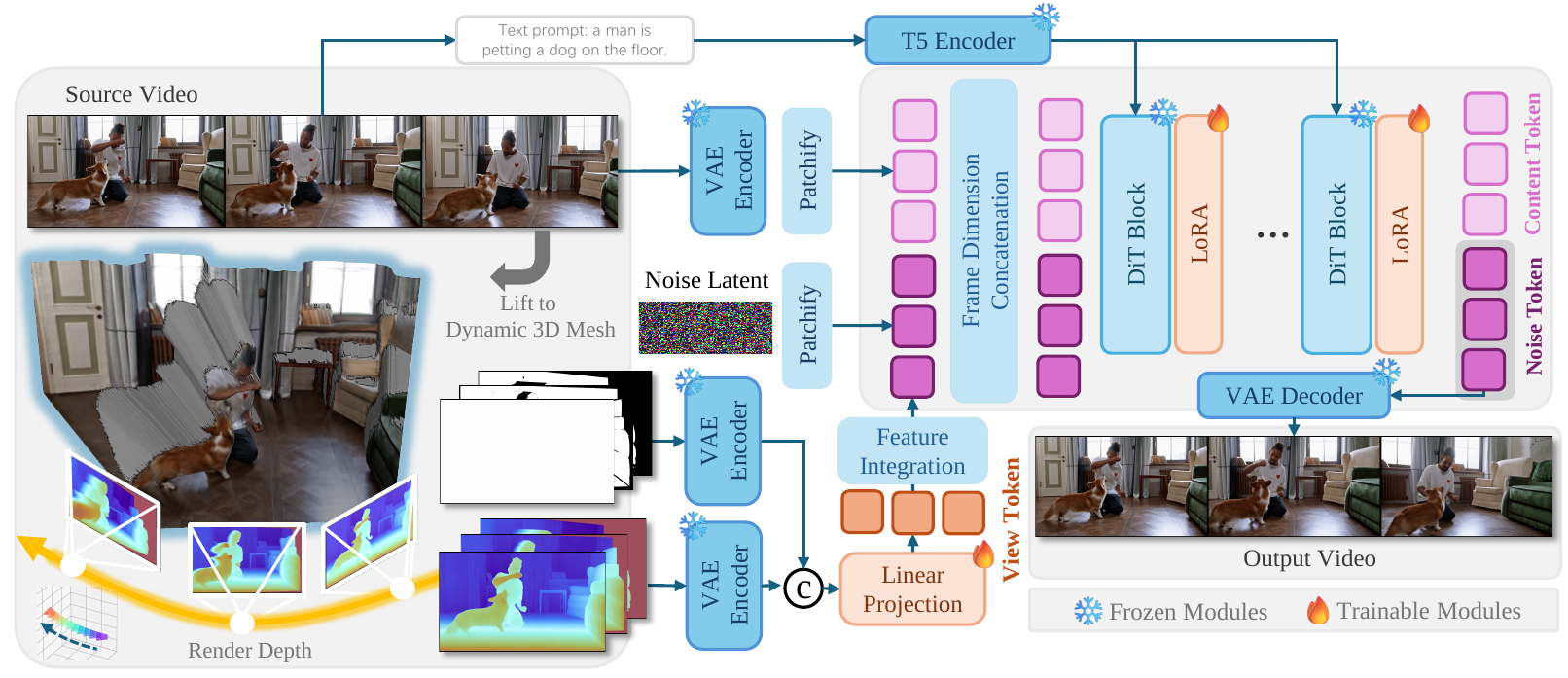}
    \caption{\textbf{Architecture overview of DepthDirector}. We render depth video and occlusion mask video under target camera viewpoint from an explicit 3D mesh, injecting it into the noise latent by projection and addition. Source video is frame-wise concatenated alongside to provide content reference. Then a LoRA-based video diffusion adapter is trained to generate video following novel camera trajectories. }
    \label{fig:overview}
    % \vspace{-10pt}
\end{figure*}

As shown in Fig.\ref{fig:overview}, the goal of our DepthDirector framework is to generate novel-view videos $V_T = \{O_t\}_{t=1}^{N}$ from an input monocular video $V_S = \{I_t\}_{t=1}^{N}$ and a target camera trajectory $\{P_t\}_{t=1}^{N}$. 
It consists of four key steps: (1) Constructing an explicit dynamic 3D mesh representation and rendering depth of novel views from it. (2) Injecting both input source video and warped depth video as generating conditions through a dual-stream conditional mechanism. (3) Building a multi-camera synchronized video dataset. (4) Using a lightweight video diffusion adapter to train our model.

\subsection{Preliminary: Text-to-Video Base Model}
\label{sec:preliminary}

Our study is conducted over Wan 2.2 pre-trained text/image-to-video foundation model\cite{wan2.2}. This architecture comprises a $3$D Variational Auto-Encoder (VAE)~\cite{kingma2022vae} for latent space mapping and a series of transformer blocks for sequence modeling. Each basic transformer block consists of a $3$D spatial-temporal attention, a cross-attention, and a feed-forward network (FFN). The text prompt embedding $c_\textrm{text}$ is obtained by T$5$ encoder $\varepsilon_{T5}$~\cite{raffel2023t5embedding} and injected into the model through cross-attention. 
We adopt the Rectified Flow~\cite{flow_lipman_2023} framework to train the diffusion transformer, %such that we can generate
{enabling the generation of} data sample $\boldsymbol{x}$ from {an initial} Gaussian sample $\boldsymbol{z}\in \mathcal{N}(\boldsymbol{0},\boldsymbol{I})$. 
Specifically, for a data point $\boldsymbol{x}$, we construct its noised version $\boldsymbol{x}_t$ at timestep $t$ as
\begin{equation}
\boldsymbol{x}_t = (1 - t)\boldsymbol{x} + t\boldsymbol{z}.
\end{equation}
The training loss is a simple mean squared error (MSE):
\begin{equation}
\mathcal{L}_{RF}(\theta)= \mathbb{E}_{t,\boldsymbol{x},\boldsymbol{z}}\left \| \boldsymbol{v}_{\theta}(\boldsymbol{x}_t, t, \boldsymbol{c}_I, \boldsymbol{c}_\textrm{text}) - (\boldsymbol{z}-\boldsymbol{x}) \right \|_2^2,
\end{equation}
where the velocity $\boldsymbol{v}_\theta$ is parameterized by the network $\theta$.

\subsection{Warped Depth for Camera Control}
\label{sec:depth}
To provide stable and easy-to-operate camera controllability, we first construct an explicit 3D representation for each frame. 
Given the input video $V_S = \{I_t\}_{t=1}^{N}$, we use the 3D foundation model\cite{wang2025pi3} to estimate the camera extrinsics $P_S=\{P_t\}_{t=1}^{N}$, and per-frame point map $X_S=\{X_t\}_{t=1}^{N}$. 
In order that the rendered depth includes less noisy artifacts, we transform the point cloud into a 3D mesh by connecting adjacent pixel points~\cite{ex4d}. Then we render the depth $D^r = \{d^r_t\}_{t=1}^{N}$ and occlusion mask $M^r= \{M^r_t\}_{t=1}^{N}$ of the 3D mesh under target camera trajectory $P_T=\{P_t\}_{t=1}^{N}$. The depth of the 3D mesh at a novel view contains occluded parts and out of frame regions, where the occlusion mask $M^r$ will be labeled zero. Therefore, $D^r$ and $M^r$ provide sufficient informations to represent the camera movements and target view layout. 

To inject this geometric information into the noise latent $\boldsymbol{x}_t$, the depth should be encoded into the same domain as the video. 
Therefore, we fit the rendered depth value $D^r$ into RGB domain, and the pretrained video VAE can be leveraged to encode depth video. We first normalize the raw depth values into $[0,1]$ in log space and then map them into RGB values by a predefined colormap.

\subsection{View-Content Dual-Stream Condition}

Having the colored depth video $D^r$ and masks $M^r$ from dynamic 3D mesh, as well as the source video $V_S$, we learn a conditional distribution using a conditional video diffusion model. To decompose the injection of both the viewpoint information from 3D mesh renders and the source video, we propose a dual-stream conditioning strategy, forcing the network to \textbf{implicitly fuse} both of the conditions and resolve the novel-view extrapolation from its understanding of video content and camera movements. 

The View Stream injects the warped depth sequence into the noise tokens, ensuring precise alignment with the target camera viewpoint. First, we encode $D^r$ and $M^r$ using the VAE encoder $\mathcal{E}$, concatenating them in the channel dimension, and project them into view tokens $\hat{\boldsymbol{x}_v}$ in noise latent space through a linear projector $F_c$: 
\begin{equation}
\begin{aligned}
    \hat{\boldsymbol{x}_v} &= F_{c}([\mathcal{E}(D^r),\mathcal{E}(M^r)]_{\text{channel-dim}}).
\end{aligned}
\end{equation}
Here, $[\cdot,\cdot]$ denotes the concatenation operation. Then, given noise latent tokens $\boldsymbol{x}_t$ in VAE compressed space, we integrate the view tokens $\hat{\boldsymbol{x}_v}$ with patchified noise latent tokens $\hat{\boldsymbol{x}_t}$ by addition, seamlessly incorporating geometric priors into the video synthesis pipeline: 
\begin{equation}
\begin{aligned}
    \hat{\boldsymbol{x}_t} &= \texttt{patchify}(\boldsymbol{x}_t) + \hat{\boldsymbol{x}_v}.
\end{aligned}
\end{equation}
While view stream resolves target camera viewpoint, it lacks semantic context for reconstructing scene appearance. We inject the source video to provide rich appearance details and dynamic motions. To achieve better synchronization and content consistency with the source video, we concatenate the source video tokens with the target video tokens along the frame dimension:
\begin{equation}
\begin{aligned}
    \hat{\boldsymbol{x}_i} &= [\texttt{patchify}(\boldsymbol{x}_s),\hat{\boldsymbol{x}_t}]_{\text{frame-dim}}.
\end{aligned}
\end{equation}
where $\boldsymbol{x}_s \in \mathbb{R}^{b \times f \times s \times d}$ is the VAE compressed latent of source video, $\hat{\boldsymbol{x}_i} \in \mathbb{R}^{b \times 2f \times s \times d}$ is the input of diffusion transformer. In other words, the input token number is doubled compared to the vanilla video generation process. Then self attention layers switch information among all tokens, ensuring the implicit condition fusion of both stream. This dual-stream design is critical for disentanglement: it relieves the View Stream from infering appearance, allowing it to focus purely on viewpoint control, while the Content Stream strictly preserves original content and temporal fidelity.

\begin{figure}[!t]
    \centering
    \includegraphics[width=0.75\linewidth]{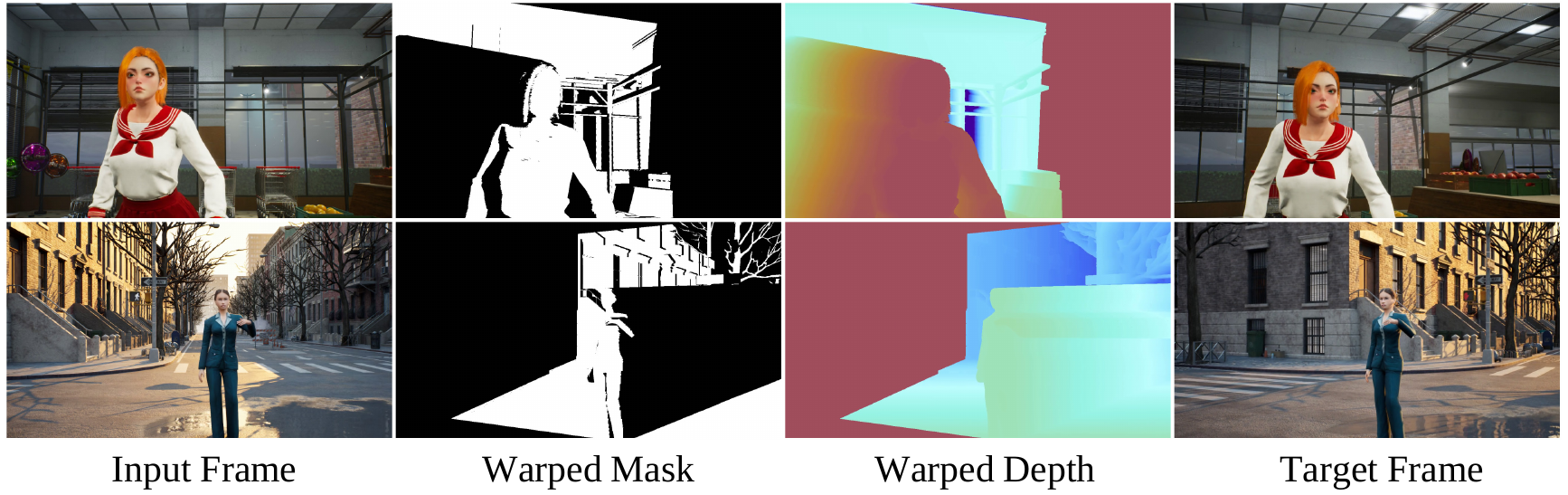}
    \caption{Illustration of our dataset: MultiCamWarp Dataset. }
    \label{fig:dataset}
    % \vspace{-15pt}
\end{figure}

\subsection{MultiCamWarp Video Dataset}
To finetune the model for our task to re-render the source video under a novel camera trajectory, we need a new paired multi-view video dataset.
%To re-render the source videos based on novel camera trajectories, paired multi-view video data is required for training. 
Specifically, the training data should include multiple shots captured in the same scene simultaneously, alongside with the ground-truth depth map to generate view condition. Acquiring such data in real-world scenarios is extremely costly, and publicly available multi-view synchronized datasets \cite{panoptic, shahroudy2016ntu, xu2024longvolcap, ego4d, ReCamMaster, camclonemaster, syncammaster} are also limited by constrained camera movements and lack of geometry information, making them unsuitable for our task. 
Therefore we chose to leverage the rendering engine to generate the training data.
% following ReCamMaster\cite{ReCamMaster}. 

We build the entire data generating system in Unreal Engine 5 \cite{unrealengine5}. Specifically, we first collect multiple 3D environments as ``backgrounds''. Then, we place animated characters within these environments as the ``main subjects'' of the videos. We then position multiple cameras facing the subjects and moving along predefined trajectories to simulate the process of simultaneous shooting. This allows us to render datasets with synchronized cameras that include dynamic objects.
To scale up the data volume, we construct a set of camera movement rules for automatically batch generation of natural and diverse camera trajectories. 
As shown in Fig.~\ref{fig:dataset}, for each dynamic scene, we render RGB value and depth value under 8 random camera trajectories, and build training data-pairs consists of warped depth video, source view video and target view video. 

Specifically, during inference, the depth scales of different input videos vary significantly. To mitigate this issue, we introduce random scaling and shifting operations before mapping the raw depth values to the RGB domain, enhancing the model’s robustness to depth-scale variations. Moreover, thanks to the dual-stream injection mechanism, the model can adaptively balance geometric and appearance information, effectively alleviating the depth distribution gap between training and inference.

Note that this pipeline can automatically build training videos on a headless Linux server with NVIDIA RTX GPU with a speed of 10 fps, making it easy to scale up. In total, we obtain 8K visually-realistic videos shot from 1K different dynamic scenes in 40 high-quality 3D environments with 8K different camera trajectories. Each video has a resolution of $720 \times 1280$ pixels and 81 frames.

\begin{figure*}[!t]
    \centering
    \includegraphics[width=1.0\linewidth]{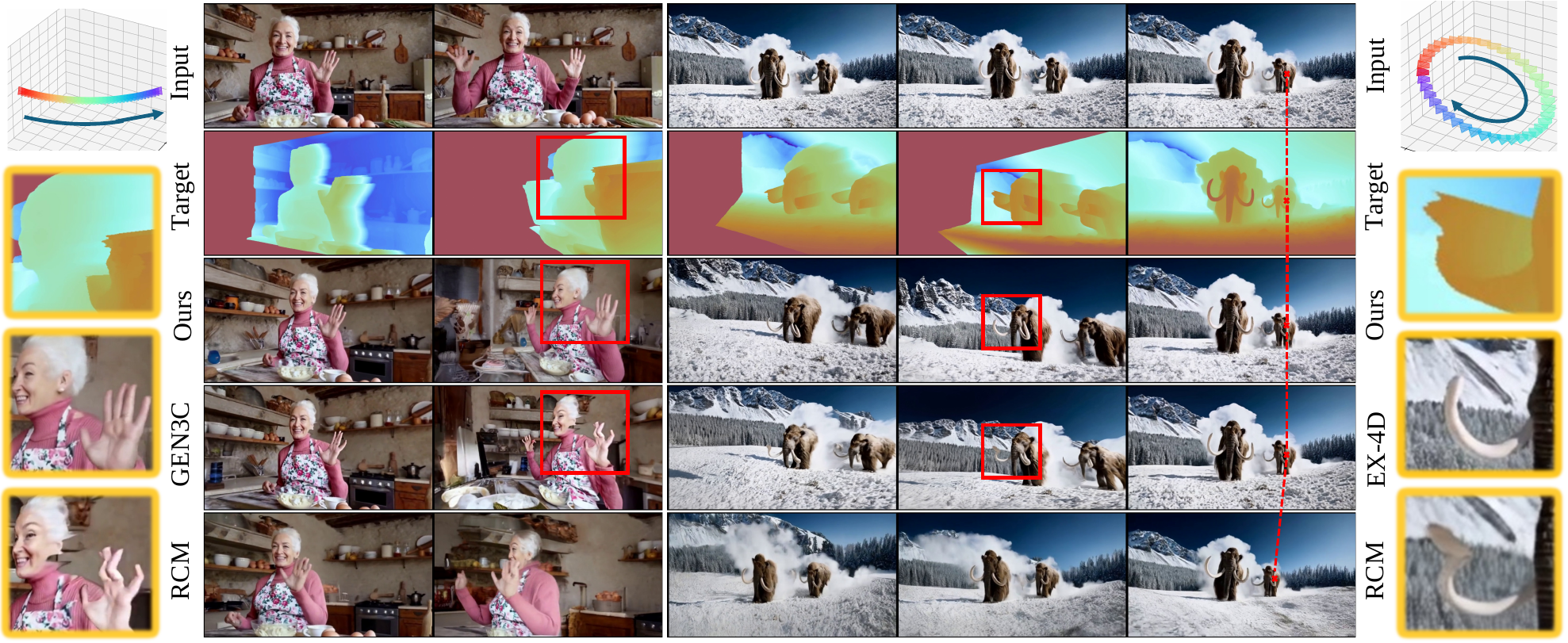}
    % \vspace{-20pt}
    \caption{\textbf{Qualitative comparison with state-of-the-art methods.} Ours DepthDirector achieves both stable camera controllability and content preservation. GEN3C\cite{ren2025gen3c} and EX-4D\cite{ex4d} generate distorted artifact due to inaccuracy of depth estimation (see zoom-out details within the yellow border). RCM(ReCamMaster)\cite{ReCamMaster} not only loses consistency at large rotation angle, but also fails to achieve stable camera control. It fails to return back to the original viewpoint in the ``orbit'' camera trajectory (see red dot on the right).}
    \label{fig:comparison1}
\end{figure*}

\subsection{Lightweight Adapter for Video Diffusion}
Instead of finetuning the entire VDM, we utilize a Low-Rank Adaptation (LoRA) \cite{lora} approach with a small rank. This allows the pre-trained VDM to remain frozen while only updating the adapter parameters. 

The training process adheres to the original flow matching objective\cite{esser2024scalingrectifiedflowtransformers}, ensuring rapid convergence and stable performance without requiring extensive computational resources. The model is trained to predict flow matching velocity, and the training objective is defined as:
\begin{equation}
\label{eq:loss}
    \mathcal{L} = \mathbb{E}_{t,\boldsymbol{z}}[\omega(t) || \boldsymbol{v}_\theta(\boldsymbol{x}_t,V_S,D^r,M^r,t;\theta)- (\boldsymbol{z}-\boldsymbol{x})||_2^2].
\end{equation}
Here $\boldsymbol{z}$, $\boldsymbol{x}$, $\boldsymbol{x}_t$, $\omega(t)$ denote the ground-truth noise, ground-truth data, noisy latents, and training weight at diffusion timestep $t$, respectively; $V_S$ is the input video; $\boldsymbol{v}_\theta$ denotes our denoising model with parameters $\theta$.

\begin{figure*}[!t]
    \centering
    \includegraphics[width=0.90\linewidth]{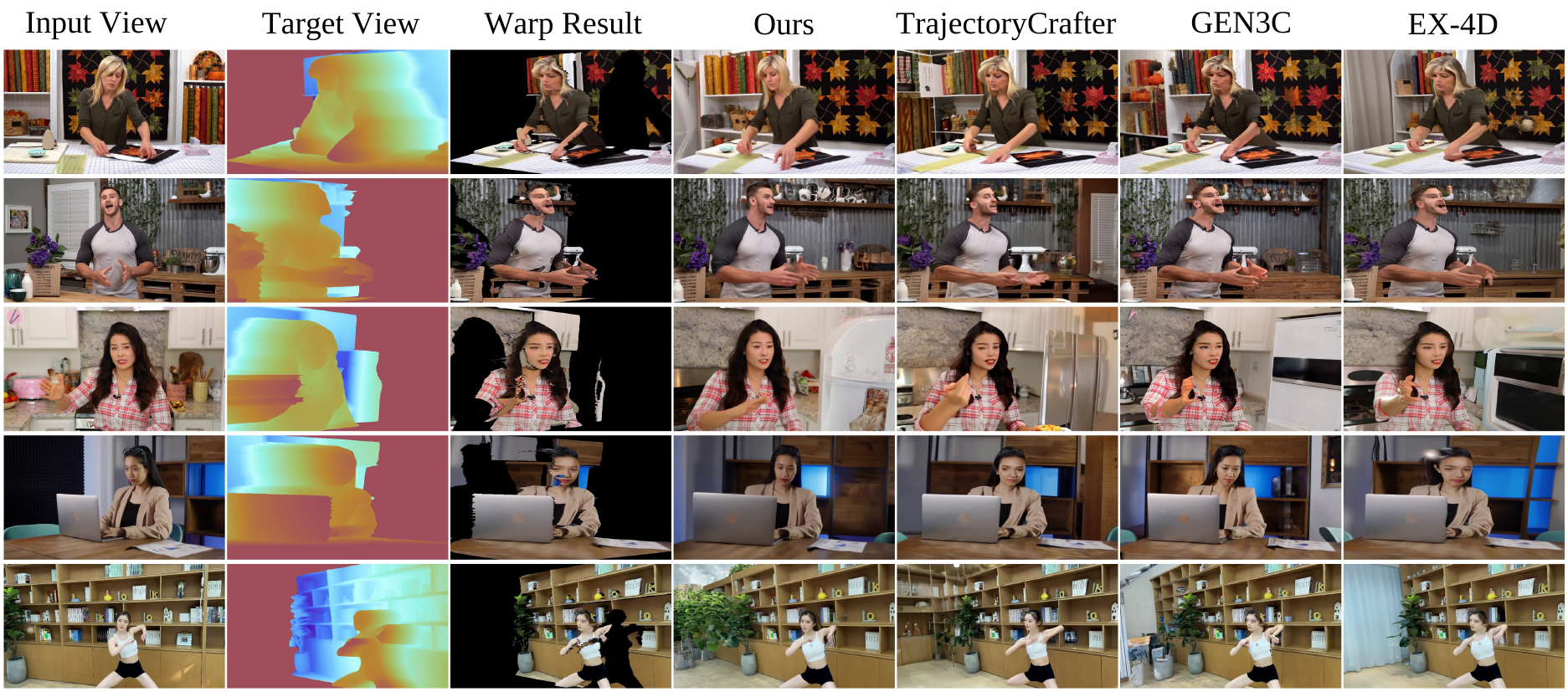}
    \caption{\textbf{Qualitative comparison with warping-based methods}. Our method preserves facial identity under camera view changes, while other methods produce noticeable distortions. }
    \label{fig:comparison_fg}
\end{figure*}

\section{Experiments}

\subsection{Implementation Details}
Our approach builds upon the Wan2.2-TI2V-5B model\cite{Wan, wan2.2}. 
The video diffusion backbone remains frozen during training, while our lightweight adapter employs a LoRA rank of 32. 
The input videos are resized to {a resolution of} $704 \times 1280$ with $81$ frames per sequence. 
The adapter is optimized using the AdamW optimizer with a learning rate of $1 \times 10^{-4}$. 
Our method is implemented on 8 NVIDIA A100 GPUs. Training is completed in 4 days, and inference generates each video in approximately 4 minutes using 50 denoising steps. Please refer to the supplementary material for additional implementation details.

\subsection{Evaluation}

\noindent\textbf{Metrics and Dataset}
For evaluation, we randomly sampled 150 in-the-wild web videos from Koala dataset\cite{koala}, which include diverse dynamic scenes and also challenging human-centric scenes. 
% Add more trajectories
To evaluate both camera controllability and novel-view synthesis capabilities, we assess our model using 10 diverse trajectories for each video. Our testing suite combines basic rotations and translations (pan, rotate, arc, zoom) with complex paths (orbit, helix, S-curve, and parabola). During these maneuvers, the camera rotates around the main subject at maximum horizontal angles between $\pm30^{\circ}$ and $\pm60^{\circ}$.

To evaluate camera trajectory accuracy, we employ the state-of-the-art camera parameters estimation model ViPE\cite{vipe} to extract camera rotation $R_i$ and translation $T_i$ for the $i$-th frame in the video. 
We then compute the Rotation Distance (RE), Translation Error (TE), and Camera Motion Consistency (CamMC), following CamI2V\cite{cami2v}. 
For content preservation evaluation, we assess human identity preservation metrics for all evaluation videos with human faces. 
We employ ArcFace\cite{arcface} embedding similarity to assess two key aspects. 
First, Reference Similarity (RS) calculates the similarity between the first frame of source video and generated frames to evaluate identity preservation. 
Second, InterFrame Similarity (IFS) quantifies the similarity between consecutive video frames to evaluate the stability of identity features during camera movements.
% To evaluate our method in terms of view synchronization\cite{ReCamMaster,camclonemaster}, we utilize the state-of-the-art image matching method GIM\cite{gim} to compute the number of matching pixels with confidence exceeding a predefined threshold, denoted as Mat. Pix.. 
% Additionally, we calculate the average CLIP similarity between source and target frames at the same timestamps, denoted as CLIP-V\cite{CVD}. 
We also evaluate our method on VBench \cite{vbench} metrics for comprehensive perceptual quality measurement, including: Subject Consistency(Cs.S.), Background Consistency(Cs.B.), Imaging Quality(Img.Q) and Motion Smoothness(M.S.).

\begin{table*}[t]
    
	\begin{center}
            % \vspace{-0.30cm}
		\caption{ \textbf{Quantitative comparison with state-of-the-art methods.} The best numbers are bolded and the second ones are unlerlined.}
            % \vspace{-0.30cm}
		\label{tab:cam_id_sync}
		\setlength\tabcolsep{3pt}
        \renewcommand{\arraystretch}{1.1}  % 增加行间距
        % \setlength\tabcolsep{0.3pt}
        % \footnotesize
        % \fontsize{8pt}{10pt}
        \scriptsize  
		\begin{tabular}{l|ccc|cc|cccc}
			\toprule
			\multirow{4}{*}{Method} & \multicolumn{3}{c|}{ \makecell[c]{Camera Accuracy}} & \multicolumn{2}{c|}{ \makecell[c]{Identity \\ Preservation}}  & \multicolumn{4}{c}{VBench}\\ 
                \cmidrule(r){2-10}
                & \makecell[c]{RE }$\downarrow$ & \makecell[c]{TE } $\downarrow$ & \makecell[c]{CamMC $\downarrow$ }
                & \makecell[c]{RS $\uparrow$} & \makecell[c]{IFS $\uparrow$} 
                & \makecell[c]{Cs.S. }$\uparrow$ & \makecell[c]{Cs.B. }$\uparrow$ & \makecell[c]{Img.Q}$\uparrow$ & \makecell[c]{M.S. }$\uparrow$
                \\
                \midrule 
                TC\cite{TrajectoryCrafter} & {1.464} & 0.133 & {2.084} & 0.567 & 0.916  & 93.85 & \underline{94.75} & 71.59 & 98.37 \\
                GEN3C\cite{ren2025gen3c} & {1.397} & \textbf{0.076} & {1.982} & 0.618 & \underline{0.943} & \underline{94.50} & 93.80 & \underline{72.04} & \underline{99.21} \\
                EX4D\cite{ex4d} & \textbf{1.186} & {0.101} & \textbf{1.687} & \underline{0.653} & 0.925 & 94.18 & 94.58 & 70.65 & 98.12 \\
                \textbf{Ours(DC)} & \underline{1.382} & \underline{0.096} & \underline{1.962} & \textbf{0.669} & \textbf{0.959} & \textbf{94.62} & \textbf{94.76} & \textbf{72.88} & \textbf{99.28} \\
                \cmidrule(r){1-10}
                RCM\cite{ReCamMaster} &  {3.576} & 0.180 & {5.063} &  {0.576} & {0.932} & 92.78 & 91.15 & 67.27 & \underline{99.17} \\
                % CamCloneMaster & 6.418 & \underline{1.027} & 9.064 & 0.5928 & 0.9426 & 507.9 & 0.9057 \\
                \textbf{Ours(1.3B)} & \underline{1.349} & \underline{0.073} & \underline{1.913} & \underline{0.653} & \underline{0.942} & \underline{93.22} & \underline{92.32} & \underline{70.00} & 98.95 \\
                % \midrule
                \textbf{Ours(Full)} & \textbf{1.010} & \textbf{0.053} & \textbf{1.434} & \textbf{0.663} & \textbf{0.954} & \textbf{94.27} & \textbf{94.46} & \textbf{72.90} & \textbf{99.21}\\
			\bottomrule
		\end{tabular}
	\end{center}
        % \vspace{-20pt}
\end{table*}

\noindent\textbf{Comparison Baselines}
For comparative evaluation, we involve the state-of-the-art methods capable of camera-controllable video synthesis, including warping-based methods\cite{TrajectoryCrafter, ex4d, ren2025gen3c}, and implicit camera conditioned methods\cite{ReCamMaster, camclonemaster}.
Other earlier methods\cite{das, trajattn, gs-dit, recapture, flovd} are omitted for their relatively inferior performance. 
%For warping-based methods, TrajectoryCrafter\cite{TrajectoryCrafter} and GEN3C\cite{ren2025gen3c} adopt 3D point cloud to warp novel view video and inject warped video with noise latent through channel dimension concatenation. EX-4D proposes to use a Depth Watertight Mesh to better handle with occlusion. 
%For implicit methods, ReCamMaster\cite{ReCamMaster} is conditioned on relative camera transformation, while CamCloneMaster\cite{camclonemaster} leverage reference video to support camera control. 
All methods are evaluated using identical camera trajectories. Note that the camera trajectories of warping-based methods are defined relative to the camera pose of each frame, a.k.a. relative camera, while those of implicit camera conditioned methods are defined relative to the first frame, a.k.a. absolute camera.
To ensure fair comparisons, we conducted our experiments under both settings. For relative camera, we follow warping-based methods to use DepthCrafter\cite{DepthCrafter} and a fixed focal length to extract per-frame depth map, named as \textbf{Ours(DC)}. And all warping-based methods receive the same depth estimation results as inputs, eliminating any advantage from differing geometric priors. 
For absolute camera, we use $\pi3$\cite{wang2025pi3} to estimate input camera parameters and per-frame point maps as in Sec.\ref{sec:depth}, named as \textbf{Ours(Full)}. 
CamCloneMaster requires reference videos as guidance, limiting the diversity of input camera trajectories. Therefore, we only evaluate arc trajectories by rendering reference videos with UE5. 

\noindent\textbf{Quantitative Comparison}
Quantitative results are presented in Table.\ref{tab:cam_id_sync}. Regarding Camera Accuracy, DepthDirector reaches achieves near-zero error, comparable to warping-based methods\cite{TrajectoryCrafter,ren2025gen3c,ex4d}. Note that the target camera trajectory is defined on the monocular point map which is inheritly inaccurate. The marginal gap in camera accuracy is introduced when DepthDirector refines this imperfect monocular geometry to prioritize visual quality. We contend that this negligible inaccuarcy on metrics is already imperceptible to users. Furthermore, under absolute camera settings, our method significantly outperforms implicit camera conditioned methods\cite{ReCamMaster}, demonstrating that DepthDirector provides stable camera controllability. 
In terms of identity preservation, DepthDirector demonstrates superior performance against all baseline methods, highlighting our method’s ability to leverage the 3D understanding of VDMs to generate results with high visual quality. 
Finally, for comprehensive perceptual quality, DepthDirector achieves the best performance on all VBench metrics. 
% This indicates that our method achieves better consistency and visual quality than baselines. 

\begin{figure}[!t]
    \centering
    \includegraphics[width=0.85\linewidth]{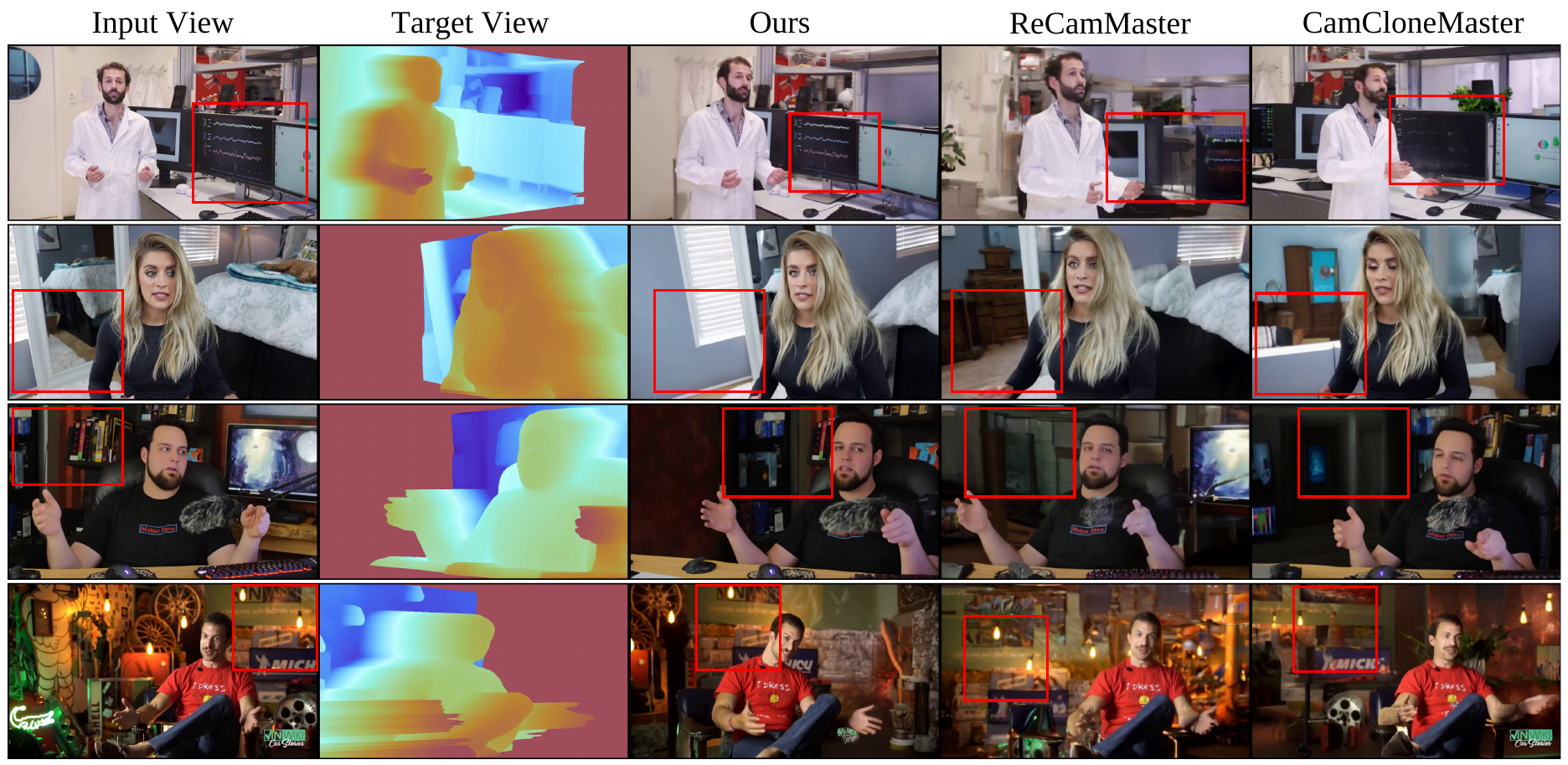}
    \caption{\textbf{Qualitative comparison with implicit-controlled methods}. Our method generates consistent background, while baseline methods fail to preserve the environment details from input views. }
    \label{fig:comparison_bk}
    % \vspace{-15pt}
\end{figure}

\noindent\textbf{Qualitative Comparison}
As shown in Fig.~\ref{fig:comparison1}, although warping-based methods can strictly follow the target trajectories, they produce distortion artifacts from inheritly inaccurate depth. Implicit methods fail to accurately align with the desired camera views and also lose consistency at large viewpoint change. Compared with them, DepthDirector not only stably aligns with target views but also generates 3D consistent content with high visual quality. This is attributed to our decomposed condition mechanism. 
% However, directly inferring camera movements for complex background remains challenging for implicit methods, because the correspondence between camera parameters and scene layout is hard to learn. 
Fig.~\ref{fig:comparison_fg} demonstrates that warping-based methods greatly suffer from inaccurate warped video, especially on human faces. However, our method can go \textbf{beyond inpainting}, leveraging its 3D understanding ability to generate 3D consistent human subject. 
As shown in Fig.~\ref{fig:comparison_bk}, DepthDirector outperforms the implicit methods on background consistency by a large margin. This is because the warped depth video provides richer geometric cues, helping the model preserve the scene geometry and appearance details.

\begin{figure}[!t]
    \centering
    \includegraphics[width=0.9\linewidth]{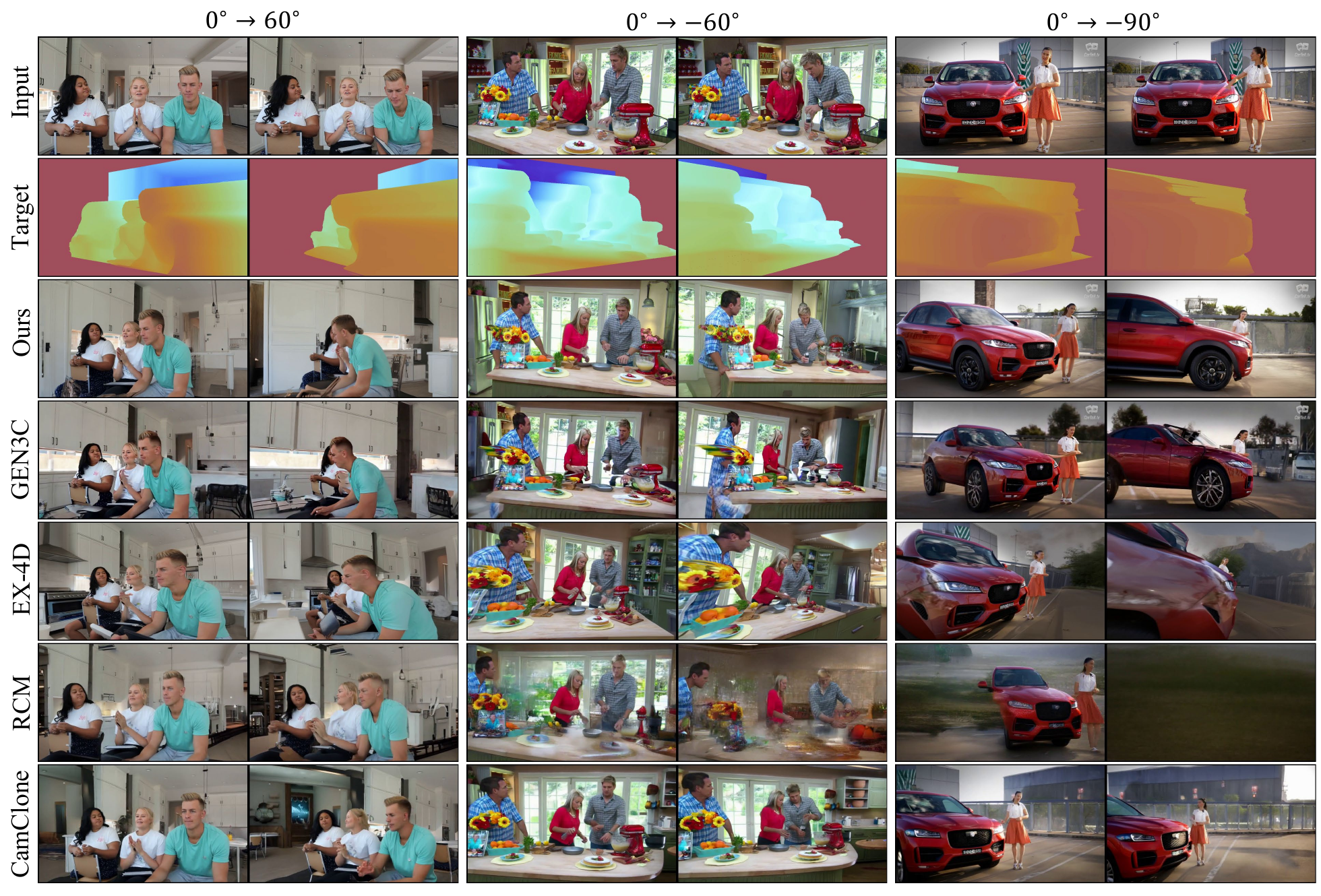}
    \caption{\textbf{Qualitative comparison on Extreme Novel View.} DepthDirector can achieve high-quality video generation results with stable camera controllability even with camera rotations of up to 90 degrees.}
    \label{fig:stress}
\end{figure}

\begin{table}[!t]
	\begin{center}
		\caption{\textbf{Quantitative comparison on Extreme Novel View.} 
        % The best numbers are bolded and the second ones are unlerlined.
        }
		\label{tab:stress}
        \setlength\tabcolsep{3pt}
        \renewcommand{\arraystretch}{1.1}  % 增加行间距
        \scriptsize  
		\begin{tabular}{l|ccc|ccc|ccc}
			\toprule
			\multirow{3}{*}{Method} & \multicolumn{3}{c|}{\makecell{Mild ($0^{\circ} \xrightarrow{} \pm30^{\circ}$)}}  & \multicolumn{3}{c|}{\makecell{Large ($0^{\circ} \xrightarrow{} \pm60^{\circ}$)}} & \multicolumn{3}{c}{\makecell{Extreme ($0^{\circ} \xrightarrow{} \pm90^{\circ}$)}} \\ 
                \cmidrule(r){2-10}
                
                & \makecell[c]{Cs.S. }$\uparrow$ & \makecell[c]{Cs.B. }$\uparrow$  & \makecell[c]{Img.Q. }$\uparrow$
                & \makecell[c]{Cs.S. }$\uparrow$ & \makecell[c]{Cs.B. }$\uparrow$ & \makecell[c]{Img.Q. }$\uparrow$
                & \makecell[c]{Cs.S. }$\uparrow$ & \makecell[c]{Cs.B. }$\uparrow$ & \makecell[c]{Img.Q. }$\uparrow$ \\
                \midrule 
                EX-4D\cite{ex4d} & 94.37 & \underline{94.58} & 70.77 & 90.09 & 91.85 & 69.43 & 86.03 & 89.02 & 67.30 \\
                GEN3C\cite{ren2025gen3c} & {94.43} & {94.05} & \textbf{72.98} & {90.97} & \underline{92.12} & \underline{70.75} & {87.73} & {90.07} & {67.64} \\
                RCM\cite{ReCamMaster} & 93.81 & 91.64 & 68.83 & 89.79 & 89.54 & 64.79 & 83.54 & 86.31 & 58.42 \\
                CamClone\cite{camclonemaster} & \textbf{95.35} & 93.63 & 70.46 & \textbf{92.63} & 91.83 & 69.99 & \textbf{90.04} & \underline{90.46} & \underline{69.32} \\
                
                % \midrule
                Ours(Full) & \underline{94.70} & \textbf{95.22} & \underline{72.92} & \underline{91.91} & \textbf{92.93} & \textbf{72.66} & \underline{89.92} & \textbf{90.91} & \textbf{71.83} \\
			\bottomrule
		\end{tabular}
	\end{center}
\end{table}

\begin{figure}[!t]
    \centering
    \includegraphics[width=1.0\linewidth]{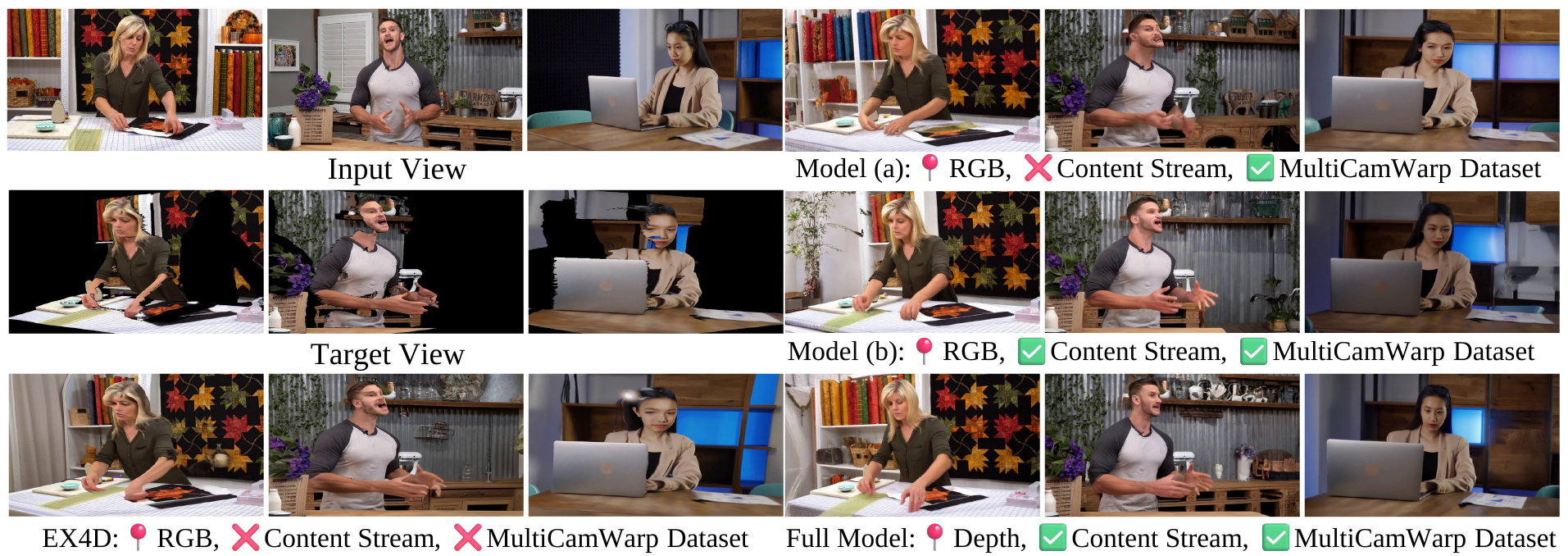}
    \caption{\textbf{Ablation Study on Model Design.} This demonstrates the importance of decomposed condition injection by depth condition and View-Content Dual-Stream mechanism. }
    \label{fig:ablation_warp}
    % \vspace{-15pt}
\end{figure}

\noindent\textbf{Comparison on Extreme Novel View}
We also performed a stress test to demonstrate the robustness of our method under extreme viewpoint change. Using VBench\cite{vbench} metrics, we assessed performance across three increasingly challenging angular ranges: Mild ($0^{\circ} \xrightarrow{} \pm30^{\circ}$), Large ($0^{\circ} \xrightarrow{} \pm60^{\circ}$), and Extreme ($0^{\circ} \xrightarrow{} \pm90^{\circ}$). We selected validation videos with fixed camera from our full evaluation set to eliminate the difference of target trajectory definition between relative and absolute camera settings. For all compared methods, the target camera originates from the input view and rotates around the main subject at specified horizontal angles. We compare against EX-4D\cite{ex4d}, GEN3C\cite{ren2025gen3c}, ReCamMaster\cite{ReCamMaster}, CamCloneMaster\cite{camclonemaster}. We use $\pi3$\cite{wang2025pi3} to extract depth map for GEN3C and our method, to demonstrate our superior performance given same geometry priors. EX-4D still utilized DepthCrafter due to its strict dependence on high-resolution depth maps. As shown in Fig.\ref{fig:stress} and Tab.\ref{tab:stress}, our method generates temporally consistent results with high fidelity and visual quality in extreme novel views, demonstrating its robust understanding of target camera movements. Notably, while CamCloneMaster achieves a higher subject consistency score, this is primarily because it fails to reach the desired camera rotation. Excluding this outlier, our method achieves highest score among all metrics. Further visualization and analysis can be found in the Supplementary Materials.

\subsection{Ablation Study}

% We carefully ablate the key components of our framework to validate their effectiveness.

\noindent\textbf{Ablation on View-Content Dual-Stream Condition}

\begin{wraptable}{r}{0.5\textwidth}
    \begin{center}
		\caption{Ablation study on Model Design. }
		\label{tab:ablation_warp}
		\setlength\tabcolsep{3pt}
        \renewcommand{\arraystretch}{1.1}  % 增加行间距
        \scriptsize  
		\begin{tabular}{l|cc|cc}
			\toprule
			\multirow{3}{*}{Method} & \multicolumn{2}{c|}{\makecell{Identity}}  & \multicolumn{2}{c}{\makecell{VBench}} \\ 
                \cmidrule(r){2-5}
                
                & \makecell[c]{RS $\uparrow$} & \makecell[c]{IFS $\uparrow$}
                & \makecell[c]{Cs.S. }$\uparrow$ & \makecell[c]{Cs.B. }$\uparrow$ \\
                \midrule 
                Model (a) & 0.5954 & 0.9606 & 94.80 & 94.57 \\
                Model (b) & \underline{0.6463} & \underline{0.9621} & \underline{95.20} & 94.43 \\
                Model (c) & 0.5804 & 0.9628 & 95.17 & \underline{94.66} \\
                
                % \midrule
                Full Model & \textbf{0.6887} & \textbf{0.9661} & \textbf{95.29} & \textbf{94.66} \\
			\bottomrule
		\end{tabular}
	\end{center}
\end{wraptable}
We extend more experiments to demonstrate how each component contributes to the goal of going beyond inpainting. We focus on metrics demonstrating content consistency, including identity preservation(RS, IFS) and VBench metrics(Cs.S, Cs.B). 
We train Model (a), which replaces the warped depth video with the warped RGB video and removes the source video concatenation in our View-Content Dual-Stream Condition mechanism, thus maintaining the same as EX-4D\cite{ex4d}. The differences between Model (a) and EX-4D are: (1) different base models, and (2) EX-4D only uses monocular video datasets to train the inpainting ability, whereas Model (a) uses a multi-camera dataset(our MultiCamWarp Dataset) for training. 
And Model (b) adds the source video concatenation based on Model (a) and still uses warped RGB video as condition. 
Model (c) adopt the same view condition stream with full model but without the source video concatenation.

\begin{figure}[!t]
    \centering
    \includegraphics[width=1.0\linewidth]{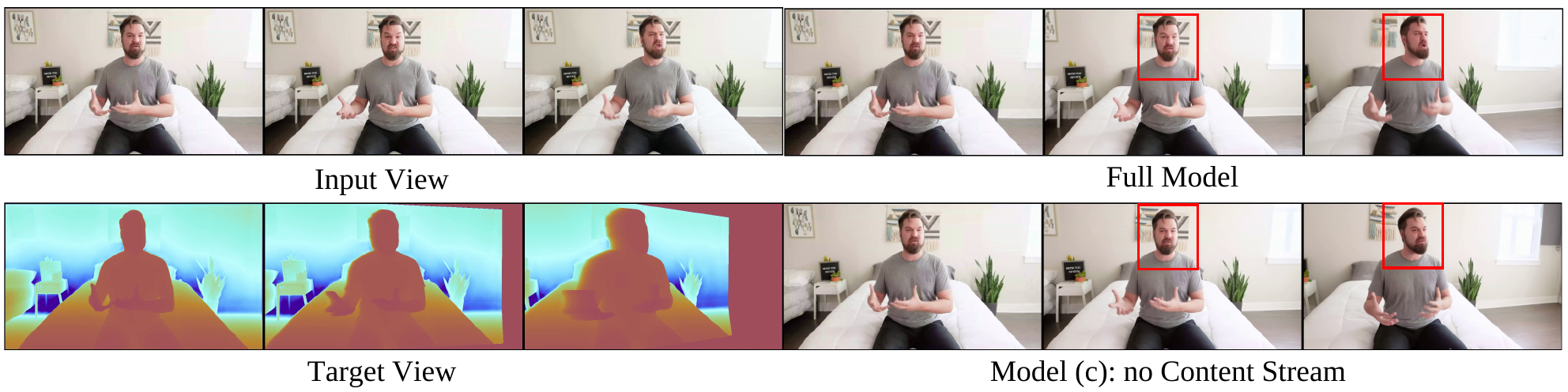}
    \caption{\textbf{Ablation Study on Model Design.} Model without injecting source video cannot recover the dynamic expression change of source video. }
    \label{fig:ablation_source}
    % \vspace{-15pt}
\end{figure}

As shown in Fig.\ref{fig:ablation_warp} and Table.\ref{tab:ablation_warp}, due to the effectiveness of our View-Content Dual-Stream Condition mechanism and the high-quality multi-camera information of our MultiCamWarp Dataset, the Model (a) and Model (b) both try to learn repainting and correction based on the inaccurate warped video. Training with multi-camera dataset helps Model (a) to learn repainting and correction instead of solely inpainting. The content stream of our Dual-Stream Condition (source video concatenation) further helps Model (b) to maintain content preservation, achieving better repainting results. However, the correction is still hard to learn due to the Inpainting Trap, and these models still generate unnatural artifacts and cannot fully recover the details. Compared with them, our Full Model can generate consistent content with high fidelity, demonstrating the necessity of depth condition. 
% Since our Full Model leverages depth information as viewpoint guidance, it is capable of going beyond inpainting and fully unleashing the 3D understanding ability of VDMs. 
As shown in Fig.\ref{fig:ablation_source}, model(c) without injecting source video infers video content from input image of first frame, and therefore fails to preserve detailed motions, such as facial expressions. 

\begin{figure*}[!t]
    \centering
    \includegraphics[width=1.0\linewidth]{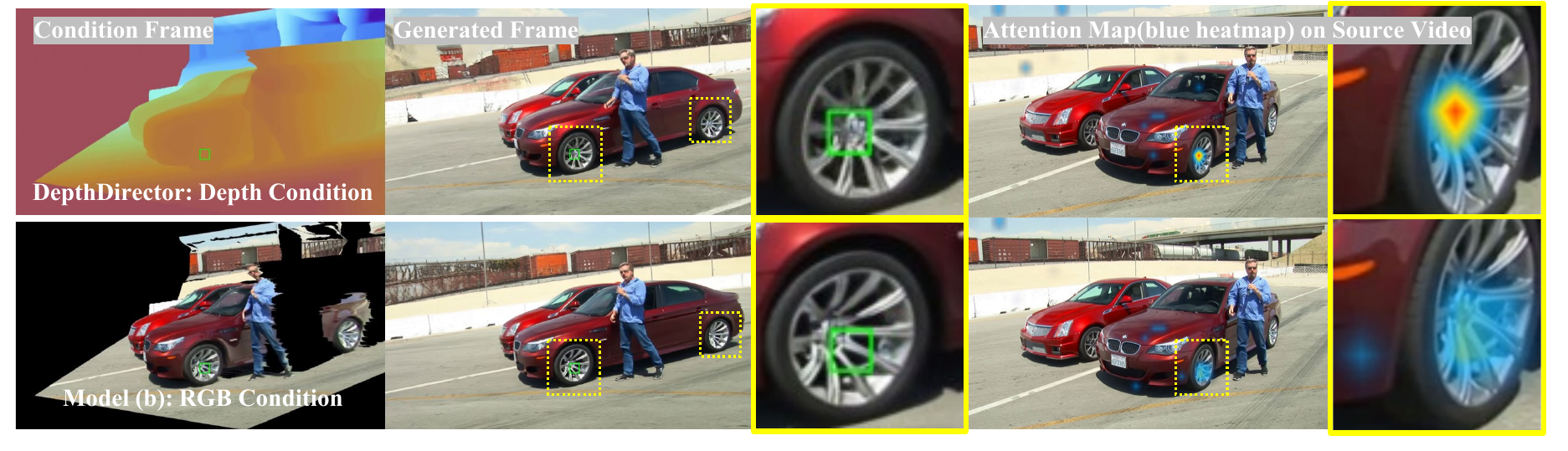}
    \caption{\textbf{Attention visualization.} Green box marks the query noise patch; the heatmap shows how source video patches attended by it. We normalize weight by a factor(0.04), red means higher. DepthDirector maintains strong and precise attentions to corresponding patch from source video, producing consistent outputs, while the RGB variant produces weak and rough attentions, leading to distorted structures. }
    \label{fig:attention}
\end{figure*}

\noindent\textbf{Mechanism Explanation}
Our core insight is to decompose the injection of video content and camera trajectories, forcing the network to infer novel views through geometric reasoning rather than inpainting. We visualize the attention weights between a noise patch(Query) and source video patches(Keys) in Fig.\ref{fig:attention}. DepthDirector precisely retrieves the corresponding patch from source video by strong attention, demonstrating the geometric reasoning ability. In contrast, Model (b) only slightly attends to a rough region. This means that the source video produces a weak contribution to the noise patch, and this weak correspondence fails to guide the model to learn consistency.

\noindent\textbf{Ablation on Base Model}
We change our base model from Wan2.2-TI2V-5B\cite{wan2.2} to Wan2.1-T2V-1.3B\cite{wan2.1}, and train at a resolution of $843\times480$. 
\textbf{Ours(1.3B)} in Table~\ref{tab:cam_id_sync} shows that changing to a lighter base model only introduces a slight performance drop. Note that, compared with ReCamMaster which is also trained on 1.3B base model, \textbf{Ours(1.3B)} still outperforms in all metrics. 

% \noindent\textbf{Ablation on Training Strategy}
% The dataset curation strategy is another key contribution of our work. In order to enhance generalizablity to various scale of inference depth video, we add random scale and shift bias adjustment to the depth video of training data-pairs. To validate its effectiveness, we conduct an ablation study to compare the novel trajectory video generation results of models trained without random depth adjustment and our full model that with random depth adjustment. The qualitative results are presented in Fig.\ref{fig:ablation_data}, 

% \begin{table}[htbp]
% \centering
% \caption{Quantitative Comparison of ID Preservation}
% \label{tab:quant_comparison}
% \begin{tabular}{lcc}
% \toprule
% Method & \multicolumn{1}{c}{RS} & \multicolumn{1}{c}{IFS} \\
% \midrule
% EX4D & 0.6760 & 0.9303 \\
% ReCamMaster & 0.5914 & 0.9394 \\
% Ours & \textbf{0.6765} & \textbf{0.9745}
% \bottomrule
% \end{tabular}
% \end{table}

\section{Conclusion}

In this paper, we propose DepthDirector, {a framework for reproducing} dynamic scenes from input videos under novel camera trajectories. 
We achieve both stable camera controlability and consistent content preservation. 
Our key idea is to leverage warped depth video as the camera-view condition, achieving decomposition of conditional injection of video content and viewpoint control. 
We design a View-Content Dual-Stream Condition mechanism that injects warped depth video to guide viewpoint changes and frame-wise concatenates source video for content reference. 
Our framework enables video diffusion models to go beyond inpainting and unleash their 3D understanding capabilities, faithfully generating videos of unseen views that are accurately aligned with the target camera trajectory. 

\section*{Acknowledgements}
This work was supported by Fundamental and Interdisciplinary Disciplines Breakthrough Plan of the Ministry of Education of China (No. JYB2025XDXM101), National Natural Science Foundation of China (No. 62220106003, 62572267) and Tsinghua University Initiative Scientific Research Program. We also thank Xiao-Lei Li, Dr. Hangtao Feng and Dr. Wensen Feng for valuable discussions.

\bibliographystyle{splncs04}
\bibliography{main}

\clearpage
\setcounter{page}{1}
\setcounter{section}{0}

\setcounter{figure}{0}
\renewcommand{\thefigure}{B\arabic{figure}}

\setcounter{table}{0}
\renewcommand{\thetable}{A\arabic{table}}

% \maketitlesupplementary

\section*{A. Details of Data Construction}

\noindent\textbf{Scene Components} We collected 40 different 3D environment assets from \url{https://www.fab.com}. To ensure data diversity, the selected scenes cover a variety of indoor and outdoor settings, such as city streets, forests, office rooms, and countryside. For the main characters, we collected 200 different human 3D models as characters and 1000 different animations to drive the collected characters. 

\noindent\textbf{Camera Trajectories}
We create camera trajectories as diverse as possible to cover various situations. We use the character’s chest position(around $150cm$ height above the ground) as the camera look-at point, and randomly sample the camera’s starting point in front of the character, ensuring the distance to the character is within the range of $[2m, 5m]$ and the pitch/yaw angles are within 10 degrees. 
From a given camera’s starting point, we selected multiple random camera trajectories. 1 to 3 points are sampled in space, and the camera moves from the initial position through these points as the movement trajectory. The total movement distance is randomly selected within the range of $[0.5,1.5]$ times of the initial distance. The rotation angles are randomly selected within 40 degrees in pitch angle and 20 degrees in yaw angle. We also shot with a static camera as the input source video. 

\noindent\textbf{Warping Construction}
For each dynamic scene, we sample 8 random camera trajectories and render the RGB value and raw depth value for each frame. We select the static camera or a random camera trajectory video as the source video, and construct the 3D mesh based on its ground-truth depth. We adopt Nvdiffrast\cite{nvdiffrast} to render the occlusion mask $M^r$ and depth $D^r$. For the rendered depth, we first clip it within the near-far range $[0.5,100]$, and then normalize it into $[0,1]$ in log space:
\begin{equation}
    D^r_i = \frac{\text{log}(D^r_i)-\text{log}(\min_{j\in S}(D^r_j))}{\text{log}(\max_{j\in S}(D^r_j))-\text{log}(\min_{j\in S}(D^r_j))}
\end{equation}
We use \verb|matplotlib.cm.get_cmap('spectral_r')| as the color-mapping to encode the raw depth value into the RGB domain. 

\begin{figure}[!t]
    \centering
    \includegraphics[width=1.0\linewidth]{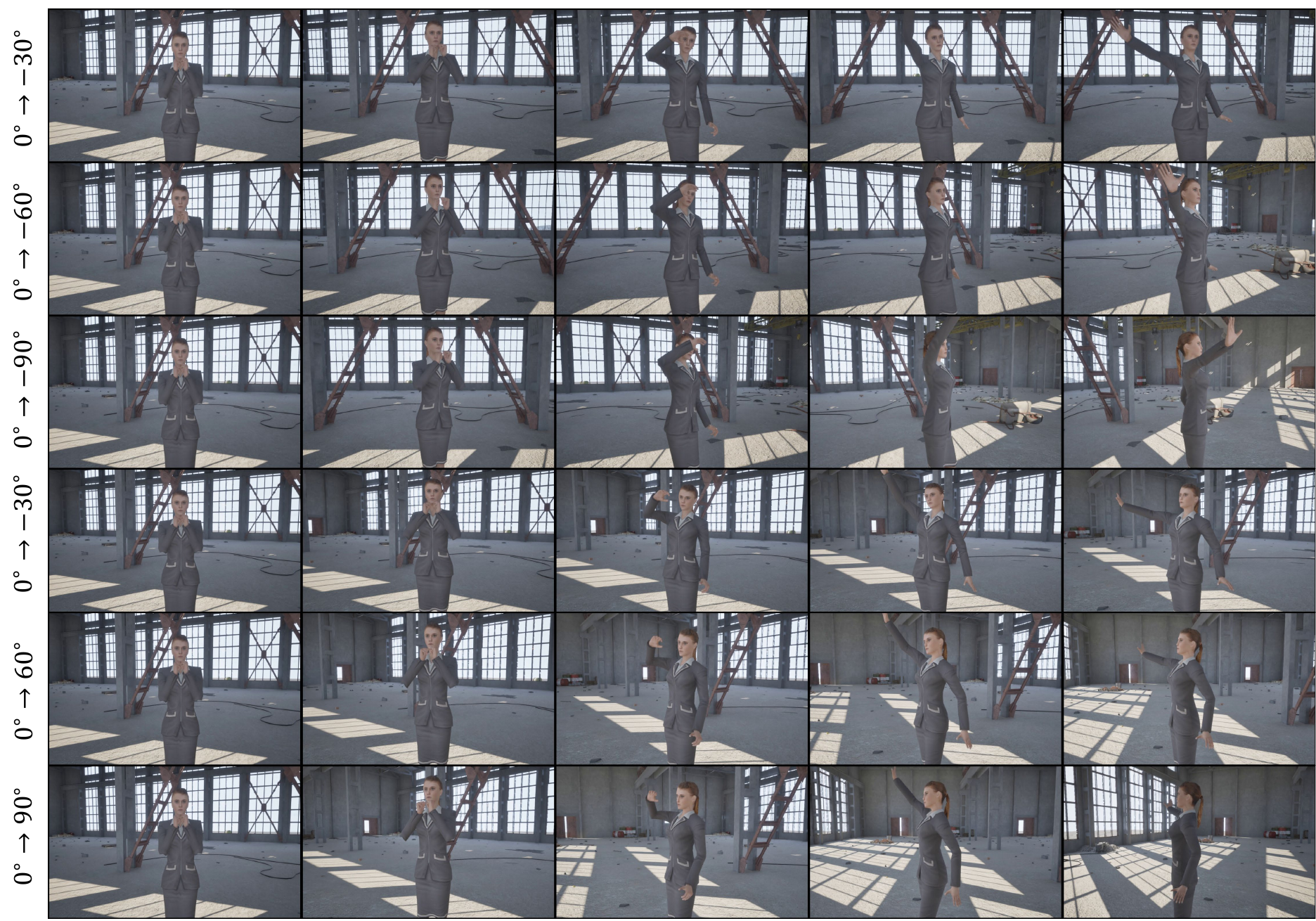}
    \caption{\textbf{Reference Video for CamCloneMaster}}
    \label{fig:camclone_ref}
\end{figure}

\section*{B. Evaluation Details}
\noindent\textbf{Baselines}
We implement all baselines following their open-source code. For a fair comparison, we use the same depth estimation method\cite{DepthCrafter} for all warping-based methods~\cite{TrajectoryCrafter,ren2025gen3c,ex4d}. We compare the impact of different depth estimation methods in Fig.\ref{fig:ablation_depth}. 
For ReCamMaster\cite{ReCamMaster}, we transform the target camera parameters into centimeters to match its scale requirement. For CamCloneMaster\cite{camclonemaster}, we use synthetic videos following the target camera trajectories as reference videos, as shown in Fig.~\ref{fig:camclone_ref}. 

\noindent\textbf{Metrics}
For RotError~\cite{cameractrl}, we evaluate per-frame camera-to-world rotation accuracy by the relative angles between ground truth rotations \(R_i\) and estimated rotations \(\tilde{R}_i\) of generated frames. We report the accumulated rotation error across all frames in radians. 
\begin{equation}
    {\rm RotErr} = \sum_{i=1}^n \cos^{-1}{\frac{\mathop{\rm tr}(\tilde{R}_i R_i^{\rm T}) - 1}{2}}    
\end{equation}
For TransError~\cite{cameractrl}, we evaluate per-frame camera trajectory accuracy by the camera location in the world coordinate system, i.e. the translation component of camera-to-world matrices. We report the sum of \({\cal L}_2\) distance between ground truth translations \(T_i\) and generated translations \(\tilde{T}_i\) for all frames.
\begin{equation}
\vspace{-1mm}
    {\rm TransErr} = \sum_{i=1}^n{\left\Vert \tilde{T}_{i} - T_{i} \right\Vert_2}
\vspace{-1mm}
\end{equation}
For CamMC~\cite{motionctrl}. We also evaluate camera pose accuracy by directly calculating \({\cal L}_2\) similarity of per-frame rotations and translations as a whole. We sum up the results of all frames.
\begin{equation}
\vspace{-1mm}
    {\rm CamMC} = \sum_{i=1}^n{\left\Vert \begin{bmatrix}\tilde{R}_i|\tilde{T}_i\end{bmatrix} - \begin{bmatrix}R_i|T_i\end{bmatrix} \right\Vert_2}
\vspace{-1mm}
\end{equation}

\section*{C. More Analysis}

\subsection*{C.1 Different Choises of Video Depth Estimation}
\label{sec:diff_depth}

\begin{figure*}[!t]
    \centering
    \includegraphics[width=1.0\linewidth]{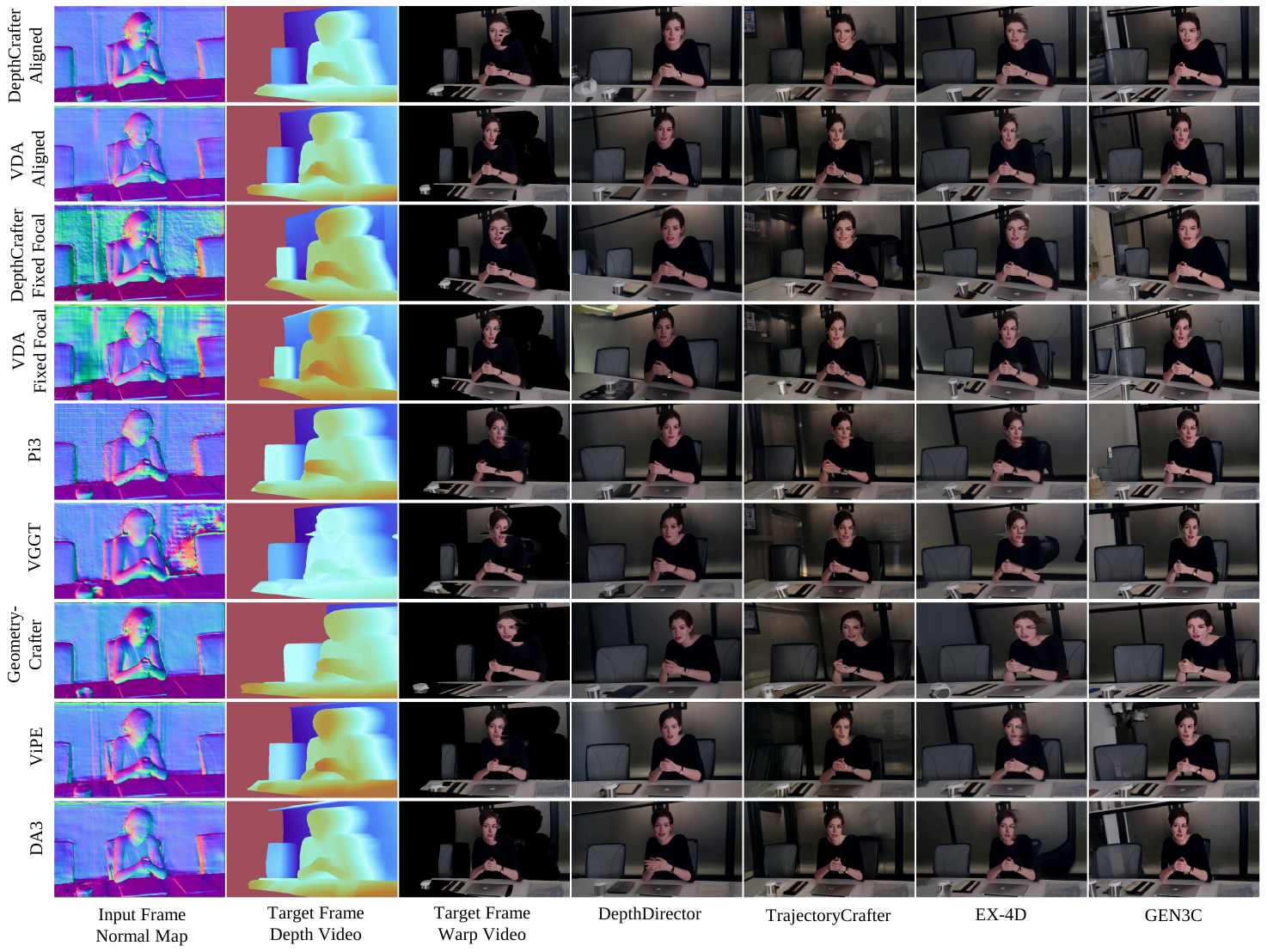}
    \caption{\textbf{Different Choises of Video Depth Estimation.} We visualize the input frame normal map derived from estimated depth map, warped depth map, warped video based on estimated depth map and the generated results of our methods and warping-based methods. Due to the inherent inaccuracy of monocular depth estimation, the warp results from these SOTA video estimation methods contain distortion and artifacts without exception. }
    \label{fig:ablation_depth}
\end{figure*}

We carefully evaluate different video depth estimation methods and investigate its impact on warping-based method\cite{TrajectoryCrafter,ren2025gen3c,ex4d} and our method. We chose the current state-of-the-art monocular video depth estimation methods: Video-Depth-Anything(VDA)\cite{VDA}, DepthCrafter\cite{DepthCrafter} and GeometryCrafter\cite{geocrafter}, alongside with 3D foundation models: VGGT\cite{wang2025vggt}, Pi3\cite{wang2025pi3}, ViPE\cite{vipe} and DA3\cite{da3}. For methods without camera intrinsics estimation, VDA and DepthCrafter, we project the depth map to point clouds with two strategies: (1) following TrajectoryCrafter~\cite{TrajectoryCrafter} and EX-4D~\cite{ex4d} to treat the depth map as inverse depth and use a fixed focal length: $\text{focal}=500$ to project into 3D space, and (2) align the depth map with Pi3~\cite{wang2025pi3}. As shown in Fig.\ref{fig:ablation_depth}, our method DepthDirector demonstrates superior robustness to different input. Due to the inherent inaccuracy of monocular depth estimation, the warp results from these SOTA video estimation methods contain distortion and artifacts without exception. Therefore, all warping-based methods generate distorted human faces.

\subsection*{C.2 Ablation Study on Dataset}

We extend more experiments to demonstrate the necessity of our proposed dataset. Another solution to generate required training data is to leverage existing MultiCam video dataset\cite{ReCamMaster}, and extract depth by off-the-shelf depth estimator. We utilized $\pi3$ to estimate the multi-cam depth at each timestamp separately. 
Due to inaccuracy of $\pi3$, warped video still misaligns with target view. 
The training results are shown in Tab.\ref{tab:dataset}. 
Note that to process a scene(10 videos) on a single 3090 GPU, $\pi3$ takes about 10 minutes, while our rendering takes only 30s, significantly saving computational resources. We will open source our rendering code to benefit the community. 

\begin{table}[t]
	\begin{center}
		\caption{Ablation Study on Dataset}
		\label{tab:dataset}
		\setlength\tabcolsep{8.3pt}

		\begin{tabular}{lcccc}
			\toprule
                {Method}  & R-Err $\downarrow$ & T-Err$\downarrow$  & Cs.Subj.$\uparrow$ & Cs.Bg.$\uparrow$ \\
                \midrule 
                
                ReCamMaster dataset & 1.88 & 0.12 & 94.96 & \textbf{94.97} \\
                \textbf{Full}(MultiCamWarp Dataset)  & \textbf{1.15} & \textbf{0.08} &  \textbf{95.03} & 94.90 \\
			\bottomrule
		\end{tabular}
    
	\end{center}
\end{table}

\begin{table}[t]
	\begin{center}
		\caption{Efficiency Comparison}
		\label{tab:efficiency}
		\setlength\tabcolsep{3pt}
        \scriptsize
		\begin{tabular}{lccccccc}
			\toprule
                {Method} & Resolution & Frames & \makecell{Denoising\\ Steps} & \makecell{ Inference\\ Time} & Base Model & \makecell{Inference\\ Time(Base)} \\
            \midrule 
                
                EX-4D & $480\times832$ & 49 & 25 & 305s & Wan2.1-I2V-14B-480P  & 250s \\
                TrajectoryCrafter & $384\times672$ & 49 & 50 & 160s & CogVideoX-Fun-5B  & 150s \\
                GEN3C & $704\times1280$ & 121 & 35 & 840s  & Cosmos-7B-Video2World & 780s \\
                ReCamMaster & $480\times832$  & 81 & 50 & 518s & Wan2.1-T2V-1.3B  & 181s \\
                CamCloneMaster & $480\times832$  & 81 & 50  & 1126s & Wan2.1-T2V-1.3B  & 181s \\
            \midrule 
                Ours(DepthDirector) & $704\times1280$ & 81 & 50  &  240s & Wan2.2-TI2V-5B & 180s \\
			\bottomrule
		\end{tabular}
    
	\end{center}
\end{table}

\begin{table}
	\begin{center}
		\caption{Quantitative Results of DaS\cite{das}}
		\label{tab:das}
		\setlength\tabcolsep{4pt}
        \scalebox{1.0}{

		\begin{tabular}{lccccccccc}
			\toprule
                {Method}  & RE & TE & CamMC & RS & IFS & Cs.S. & Cs.B. & Img.Q & M.S.  \\
                \midrule 
                DaS\cite{das} & 3.806 & 0.294 & 5.095 & 0.564 & 0.916 & 92.37 & 92.58 & 67.84 & 98.81  \\
			\bottomrule
		\end{tabular}
        }
	\end{center}
\end{table}

\subsection*{C.3 Efficiency Comparison}

We conduct an efficiency comparison with baseline methods on a single NVIDIA A100 GPU. Results are shown in Tab.\ref{tab:efficiency}.

\subsection*{C.4 Extreme Novel View}

\begin{figure*}[!t]
    \centering
    \includegraphics[width=1.0\linewidth]{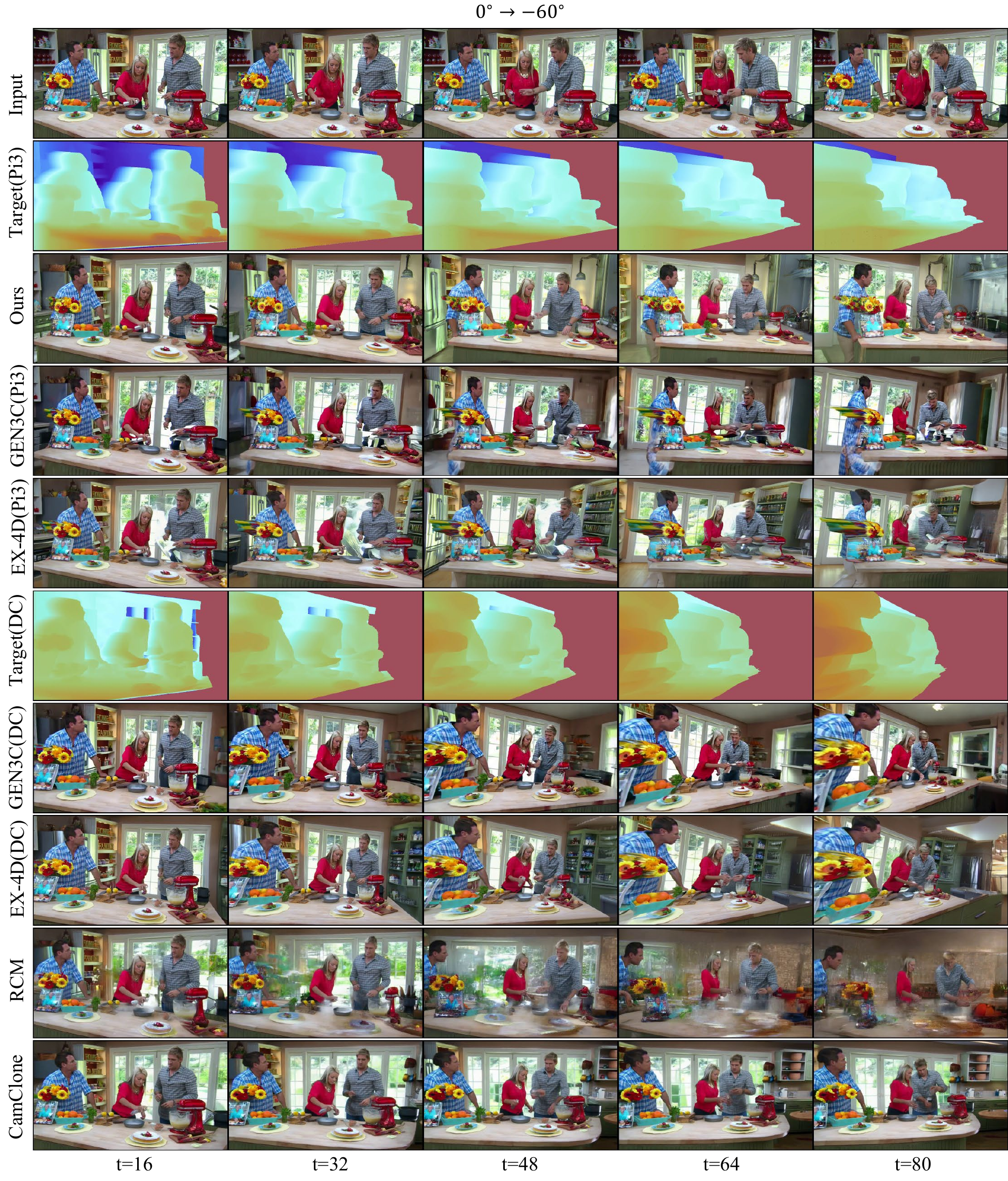}
    \caption{\textbf{More Comparison of Extreme Novel View. }}
    \label{fig:stress_1}
\end{figure*}

\begin{figure*}[!t]
    \centering
    \includegraphics[width=1.0\linewidth]{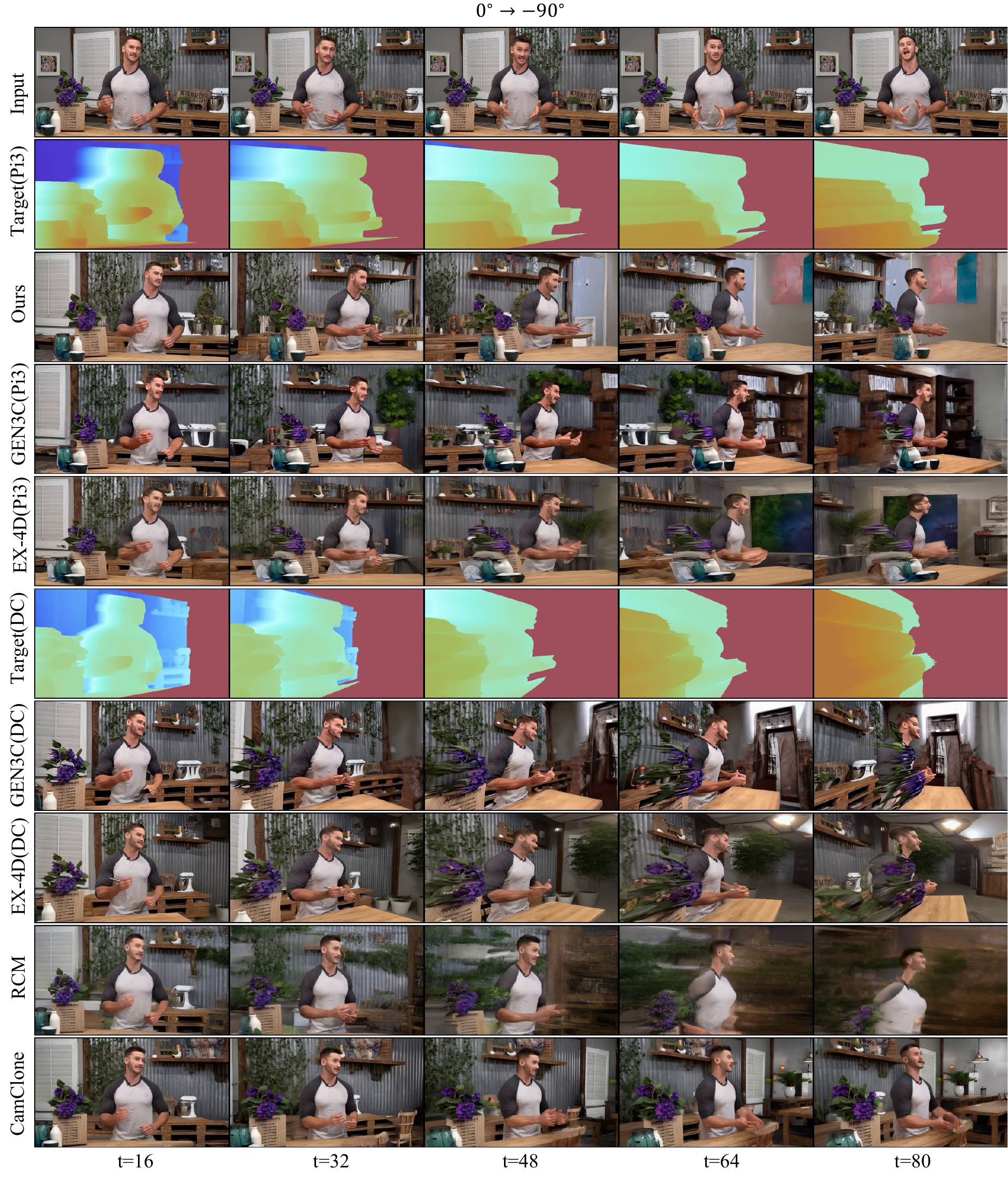}
    \caption{\textbf{More Comparison of Extreme Novel View. }}
    \label{fig:stress_2}
\end{figure*}

\begin{figure*}[!t]
    \centering
    \includegraphics[width=1.0\linewidth]{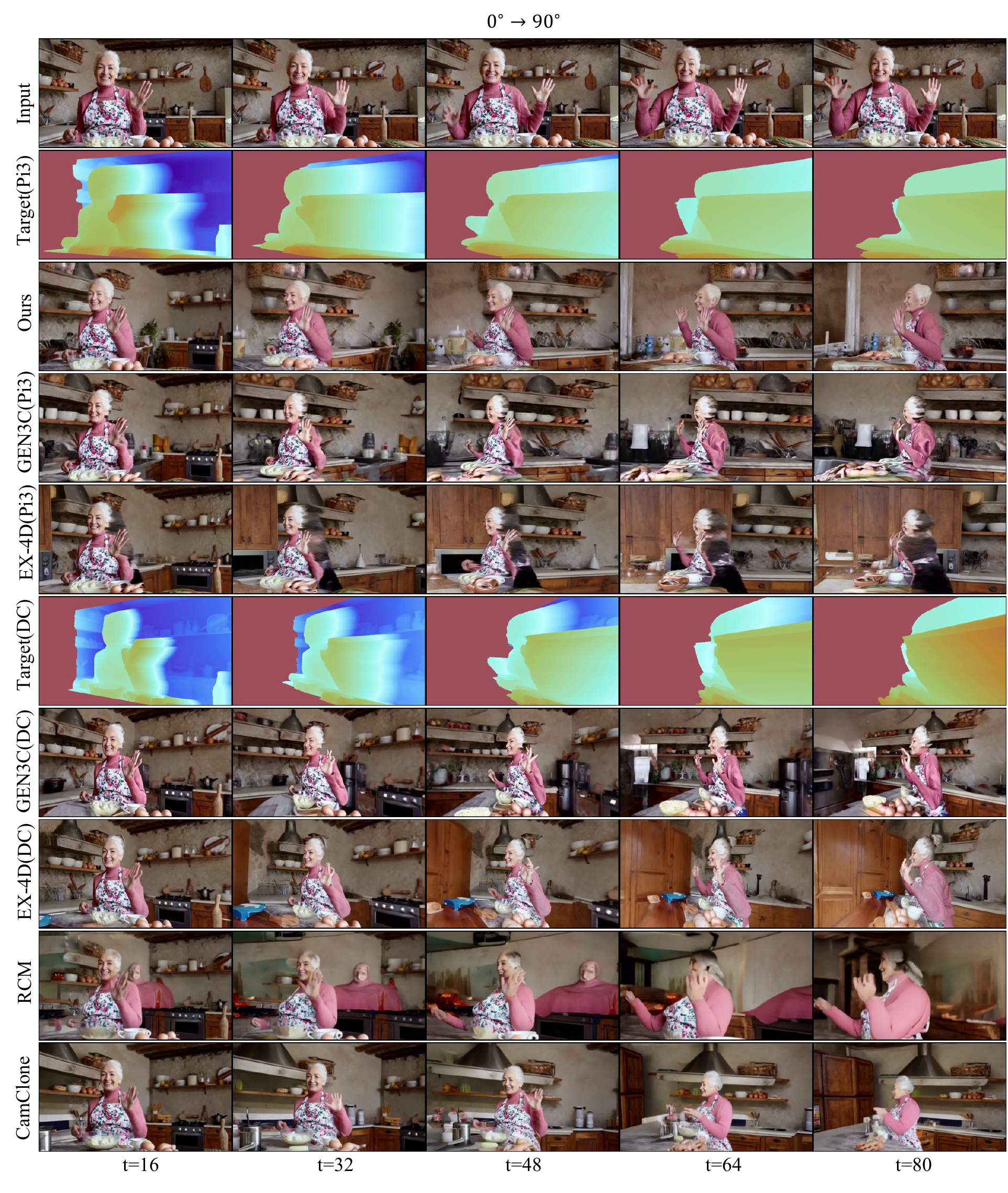}
    \caption{\textbf{More Comparison of Extreme Novel View. }}
    \label{fig:stress_3}
\end{figure*}

\begin{figure*}[!t]
    \centering
    \includegraphics[width=1.0\linewidth]{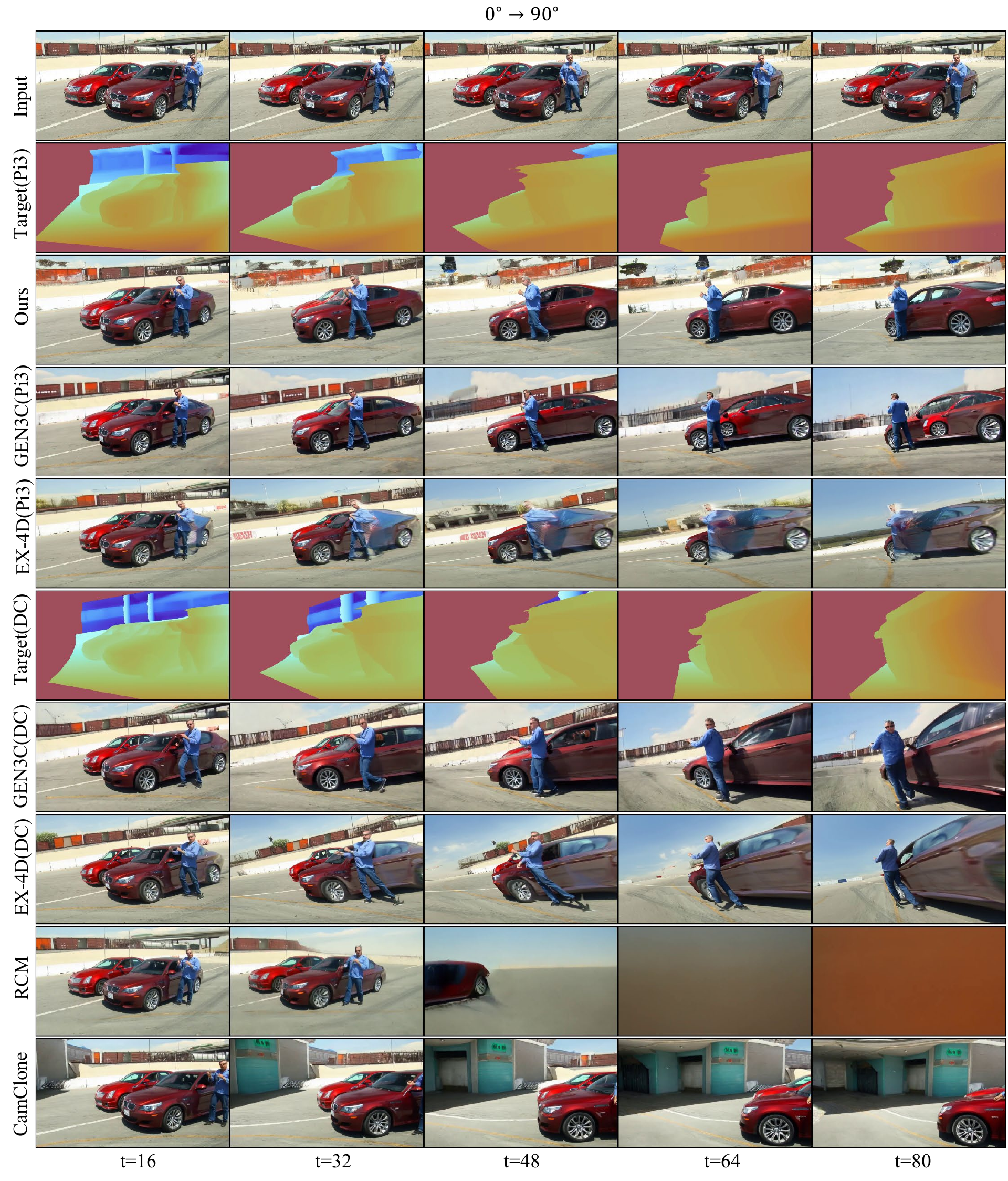}
    \caption{\textbf{More Comparison of Extreme Novel View. }}
    \label{fig:stress_4}
\end{figure*}

We compare Camera Accuracy metrics with implicit-based methods on extreme novel view evaluation. As shown in Tab.\ref{tab:stress_camera}, our method outperforms all baselines. When rotation angles increase, our method still achieves a constant near zero translation error, demonstrating the robustness of camera controllability. Note that the rotation errors grow linearly with the target rotation angles, which is a natural bias.  More visualizations on extreme novel view are shown in Fig.\ref{fig:stress_1}, Fig.\ref{fig:stress_2}, Fig.\ref{fig:stress_3}, Fig.\ref{fig:stress_4}.

\subsection*{C.4 Robustness to Depth Noise}

\begin{figure*}[!t]
    \centering
    \includegraphics[width=1.0\linewidth]{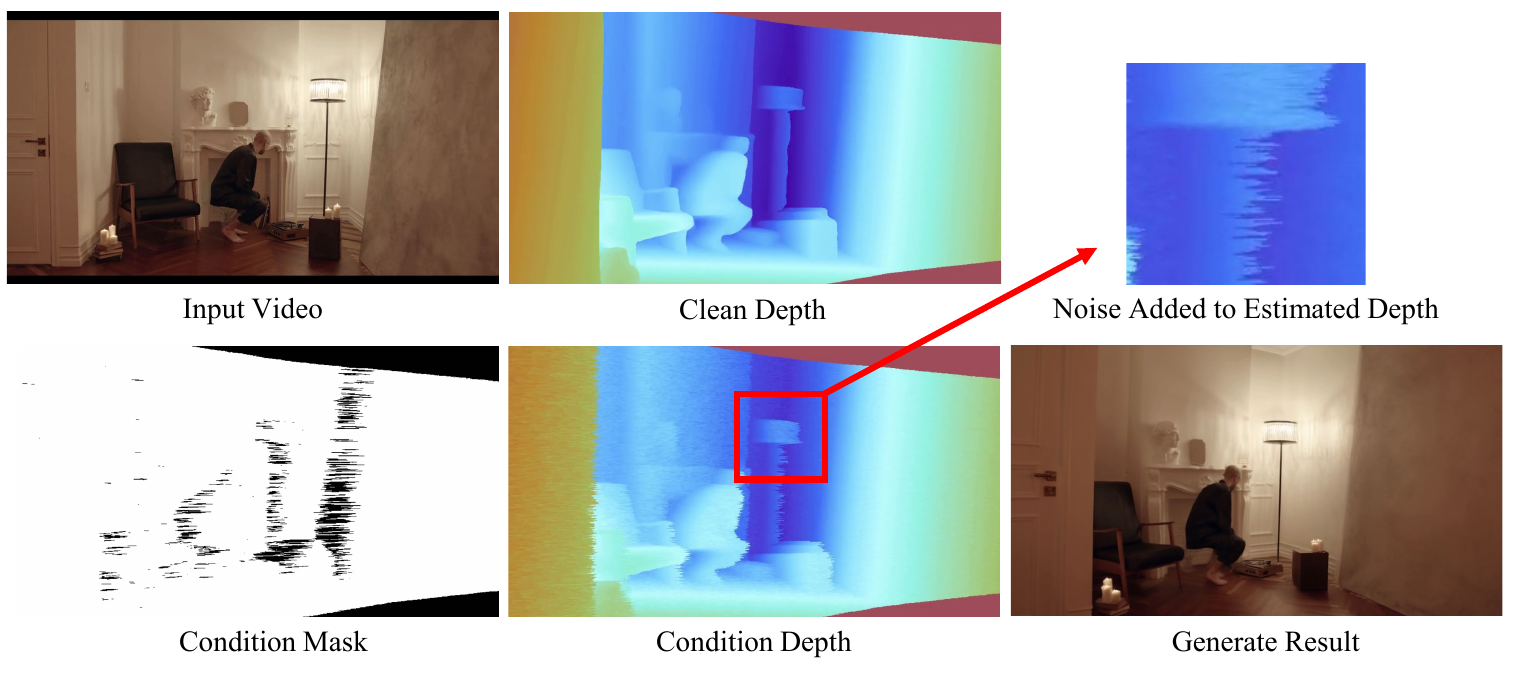}
    \caption{\textbf{Robustness.} We add Gaussian noise (std $= 10\%$ of mean depth) to the estimated depth map before warping the geometry into novel camera trajectories. The generated video still remains temporally consistent and follows the target camera viewpoints. }
    \label{fig:depth_noise}
\end{figure*}

We additionally conduct an ablation on depth noise levels. As shown in Fig.\ref{fig:depth_noise}, even under Gaussian noise that corrupts thin structures, DepthDirector remains robust. This is attributed to the effectiveness of the decomposed injection of video content and camera trajectories. 

\subsection*{C.5 Comparison with DaS\cite{das}}

While both DaS and ours avoid RGB, the principle differs. 
DaS conditions on first frame image and 3D tracking, latter encodes \emph{both} camera \emph{and} object dynamics. It means heavy dependence on this condition to infer dynamic content, replacing one form of content leakage with another.
In contrast, our key contribution is the \emph{decomposition} of camera and content signals, not merely the removal of RGB.
Furthermore, 3D tracking relies on point tracking estimators, introducing more inference artifacts. As shown in Tab.\ref{tab:das}, DaS performs substantially worse than all baselines, confirming that it's not competitive in rerendering task.

\section*{D. More Results}

\begin{figure*}[!t]
    \centering
    \includegraphics[width=1.0\linewidth]{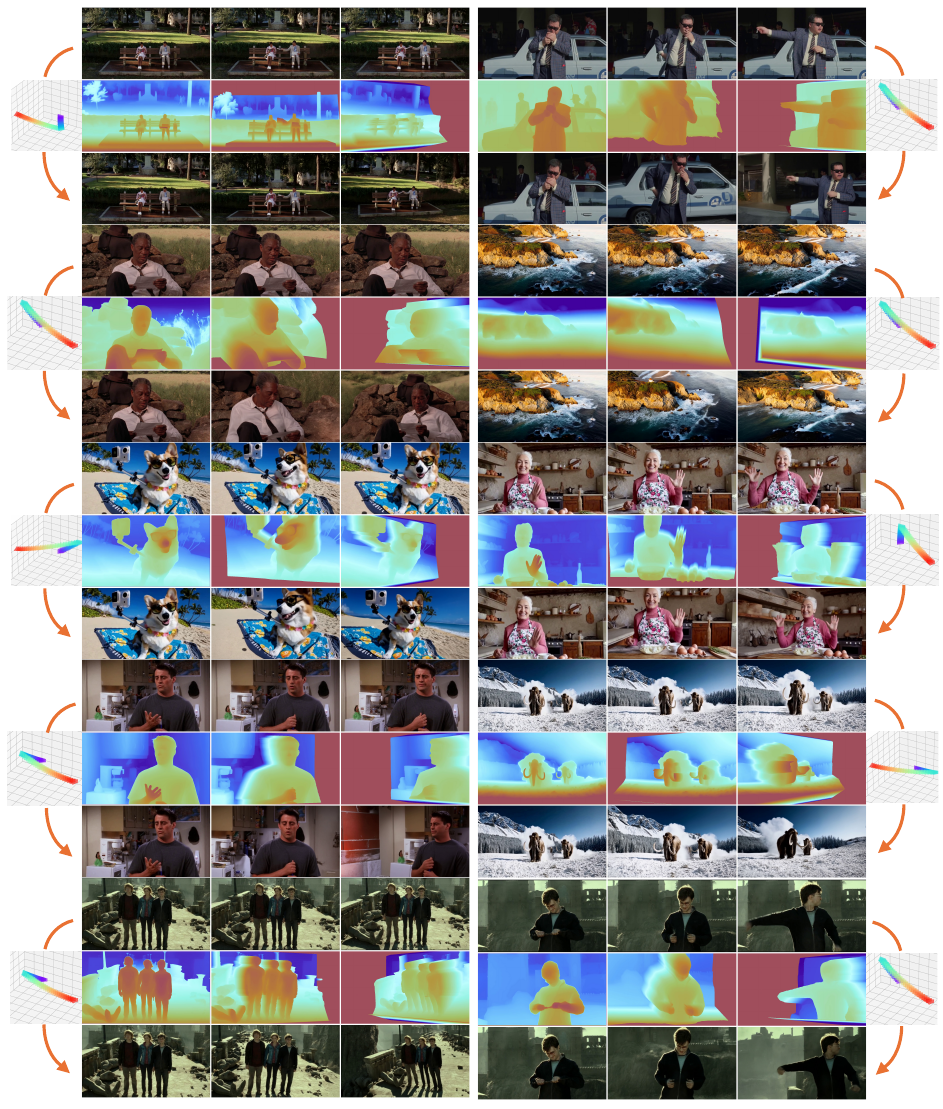}
    \caption{\textbf{More Results of our method. }}
    \label{fig:more_results}
\end{figure*}

\begin{figure*}[!t]
    \centering
    \includegraphics[width=1.0\linewidth]{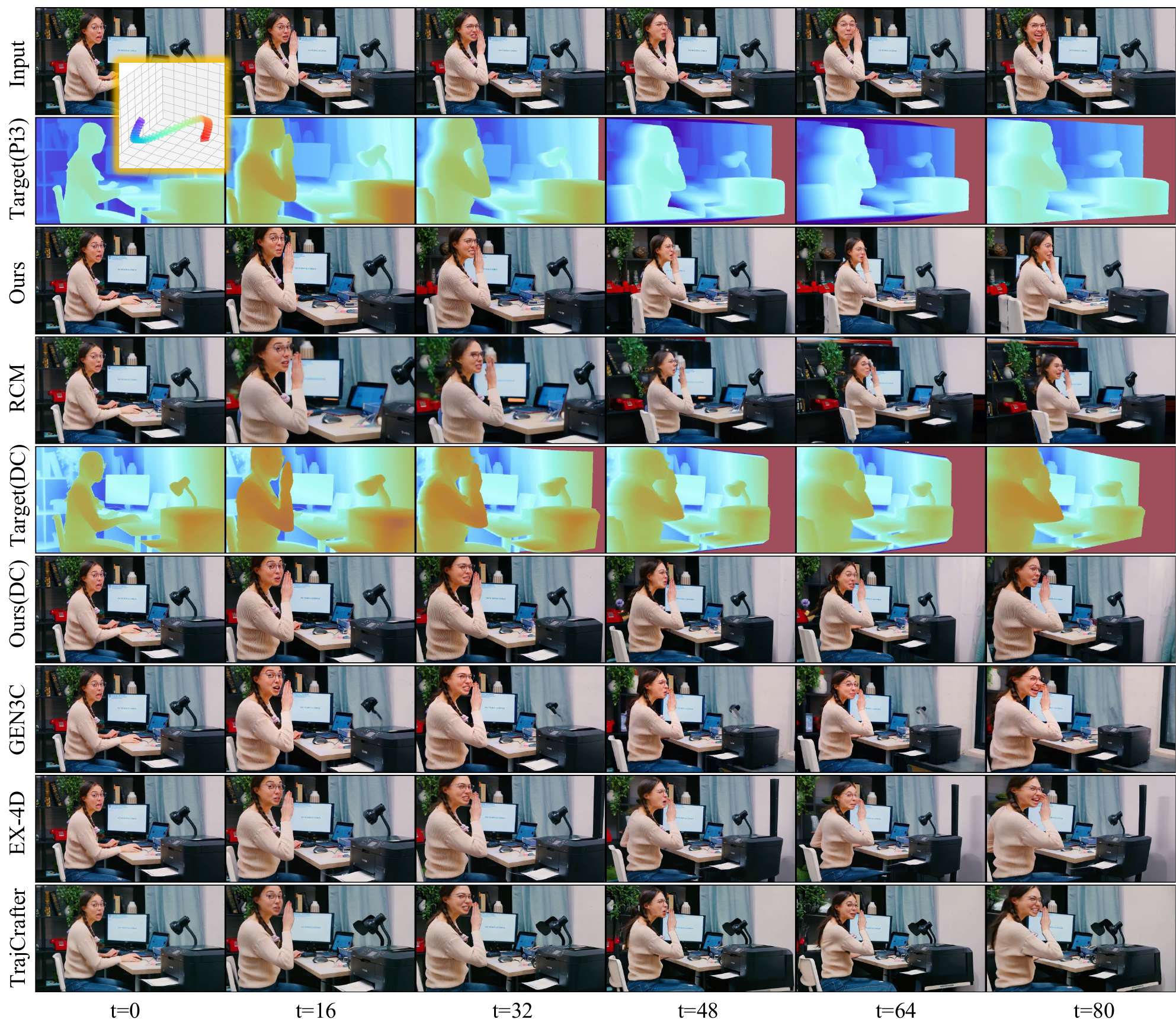}
    \caption{\textbf{More Comparison of our method. }}
    \label{fig:more_1}
\end{figure*}

\begin{figure*}[!t]
    \centering
    \includegraphics[width=1.0\linewidth]{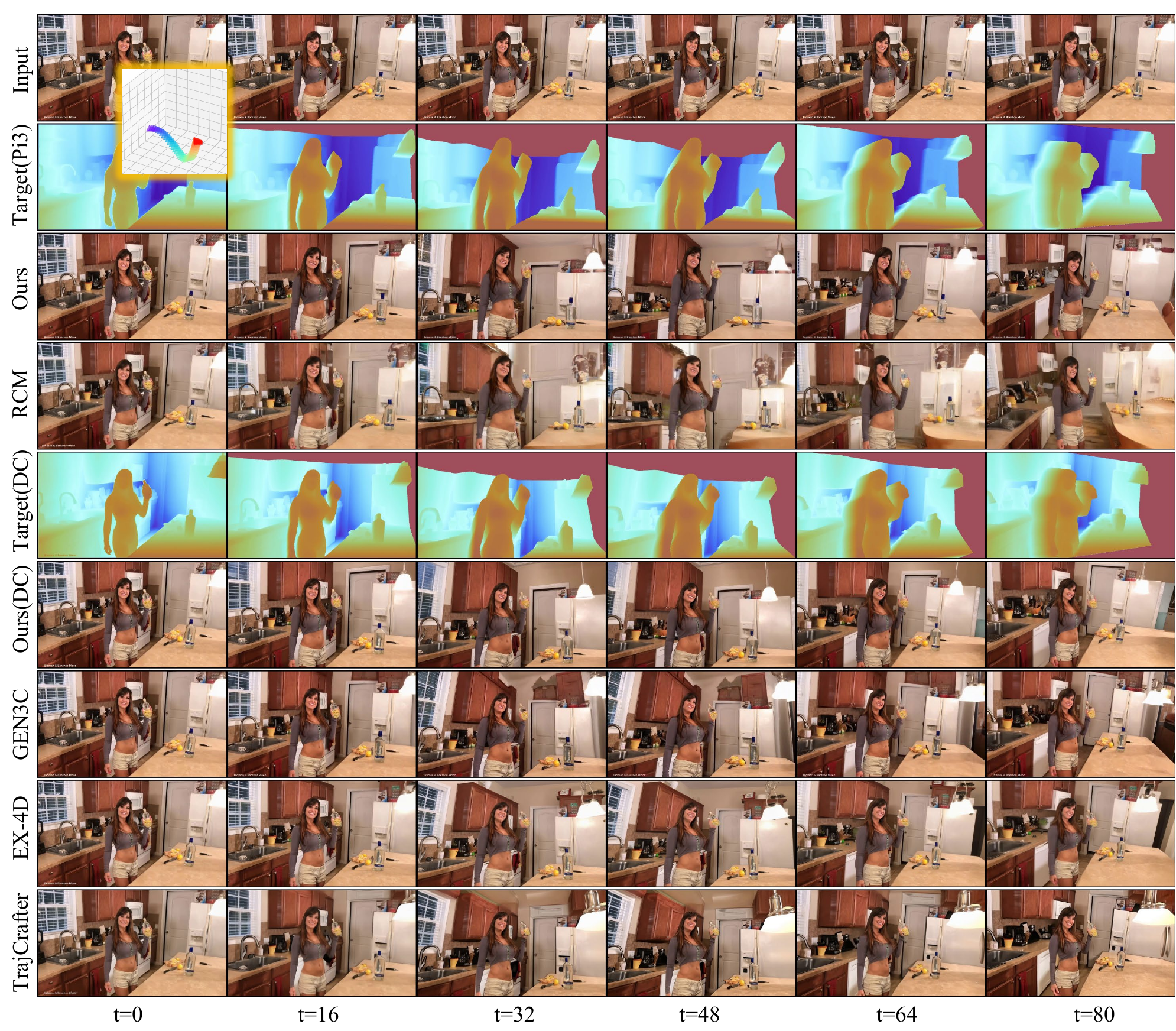}
    \caption{\textbf{More Comparison of our method. }}
    \label{fig:more_2}
\end{figure*}

\begin{figure*}[!t]
    \centering
    \includegraphics[width=1.0\linewidth]{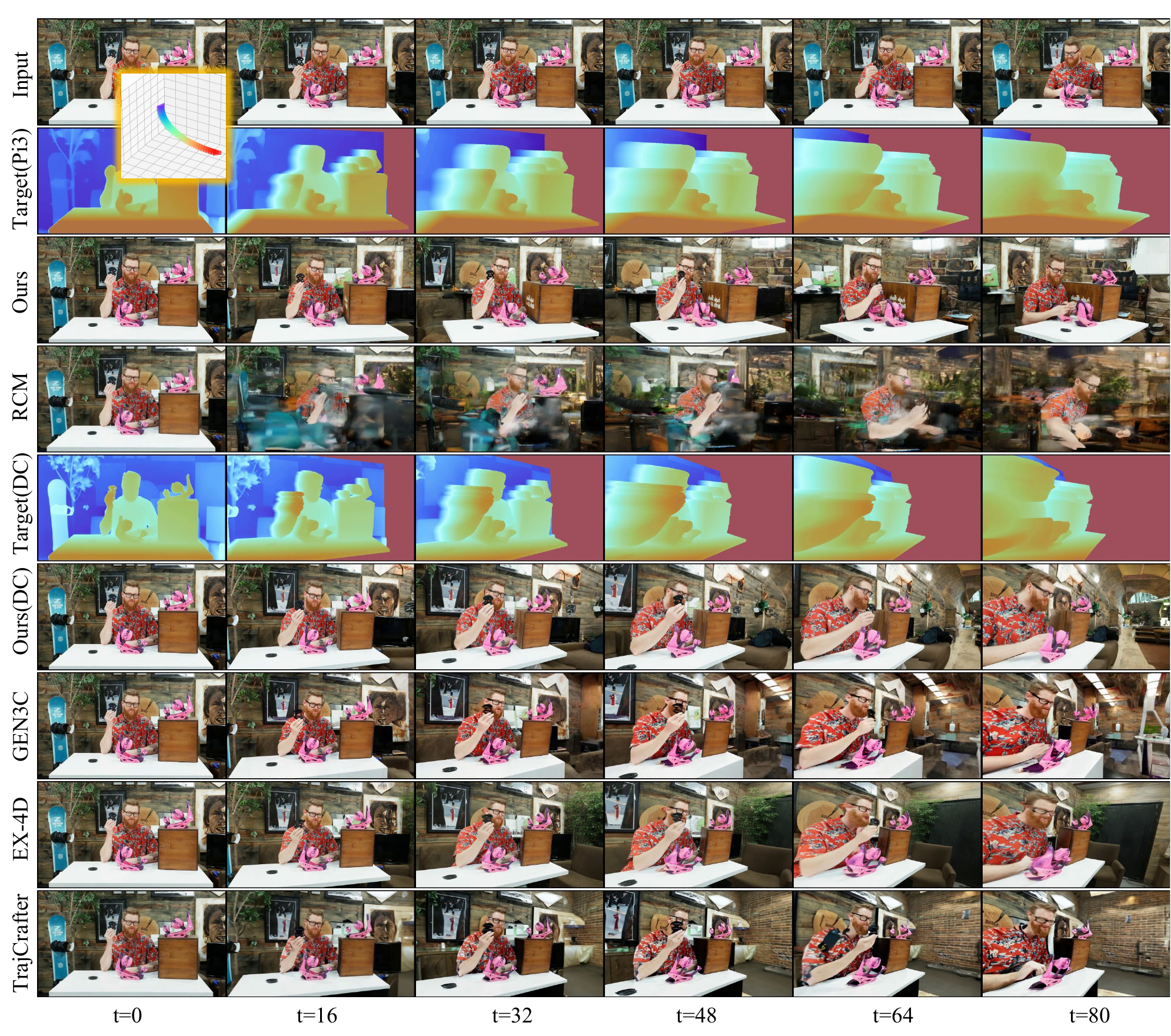}
    \caption{\textbf{More Comparison of our method. }}
    \label{fig:more_3}
\end{figure*}

\begin{figure*}[!t]
    \centering
    \includegraphics[width=1.0\linewidth]{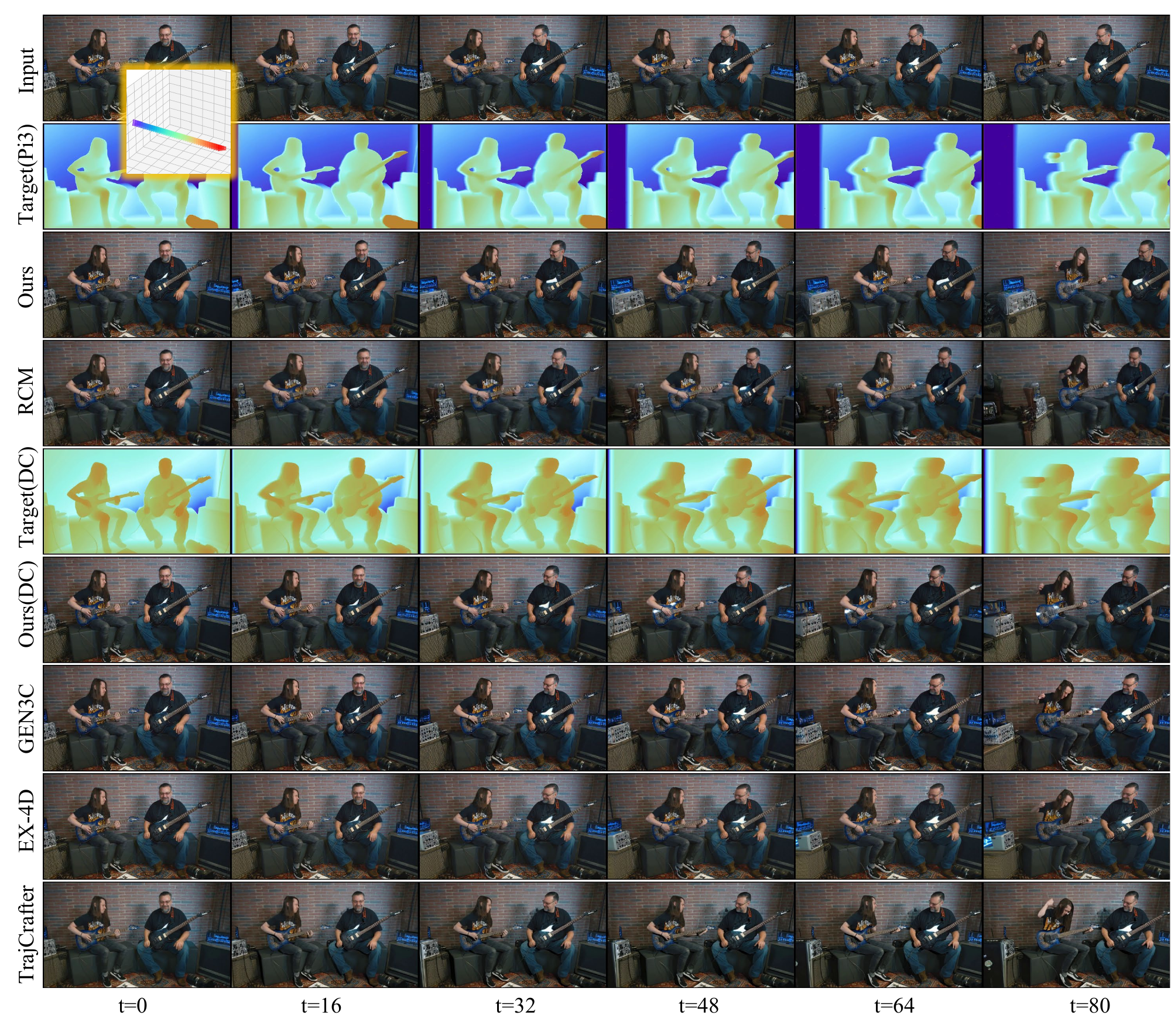}
    \caption{\textbf{More Comparison of our method. }}
    \label{fig:more_4}
\end{figure*}

In main paper, we conduct our evaluation on multiple challenging trajectories. More visualizations are shown in Fig.\ref{fig:more_1}, Fig.\ref{fig:more_2}, Fig.\ref{fig:more_3}, Fig.\ref{fig:more_4}.
And we demonstrate more synthesized results of DepthDirector, presented in Fig.\ref{fig:more_results}.

\begin{table}[!t]
	\begin{center}
		\caption{Stress Test On Extreme Viewpoint Change}
		\label{tab:stress_camera}
        \setlength\tabcolsep{3pt}
        \renewcommand{\arraystretch}{1.1}  % 增加行间距
        \scriptsize  
		\begin{tabular}{l|ccc|ccc|ccc}
			\toprule
			\multirow{3}{*}{Method} & \multicolumn{3}{c|}{\makecell{Mild ($0^{\circ} \xrightarrow{} \pm30^{\circ}$)}}  & \multicolumn{3}{c|}{\makecell{Large ($0^{\circ} \xrightarrow{} \pm60^{\circ}$)}} & \multicolumn{3}{c}{\makecell{Extreme ($0^{\circ} \xrightarrow{} \pm90^{\circ}$)}} \\ 
                \cmidrule(r){2-10}
                
                & \makecell[c]{RotErr }$\uparrow$ & \makecell[c]{TErr}$\uparrow$ & \makecell[c]{CamMC }$\uparrow$
                & \makecell[c]{RotErr }$\uparrow$ & \makecell[c]{TErr}$\uparrow$ & \makecell[c]{CamMC }$\uparrow$
                & \makecell[c]{RotErr }$\uparrow$ & \makecell[c]{TErr}$\uparrow$ & \makecell[c]{CamMC }$\uparrow$
                \\
                \midrule 
                % EX-4D\cite{ex4d} & \underline{1.900} & {0.117} & \underline{2.691} & \underline{2.908} & 0.104 & \underline{4.114} & \textbf{3.964} & 0.122 & \textbf{5.605} \\
                % GEN3C\cite{ren2025gen3c} & {2.250} & \underline{0.073} & {3.183} & {3.454} & \underline{0.0574} & {4.885} & {4.268} & \underline{0.070} & {6.034} \\
                RCM\cite{ReCamMaster} & 4.809 & 0.104 & 6.799 & 6.985 & 0.105 & 9.857 & 8.493 & 0.154 & 11.991 \\
                CamClone\cite{camclonemaster} & 5.543 & 0.174 & 7.834 & 8.556 & 0.094 & 12.077 & 11.45 & 0.093 & 16.155 \\
                
                \midrule
                Ours(Full) & \textbf{1.281} & \textbf{0.032} & \textbf{1.813} & \textbf{2.499} & \textbf{0.043} & \textbf{3.534} & \textbf{4.031} & \textbf{0.035} & \textbf{5.695} \\
			\bottomrule
		\end{tabular}
	\end{center}
\end{table}

\section*{E. Limitations}

Our framework enables video diffusion models to go beyond inpainting and unleash their 3D understanding capabilities, faithfully generating videos of unseen views that are accurately aligned with the target camera trajectory. There are nevertheless some limitations. Our method loses camera controllability for inputs lacking geometric features. For example, if the input consists only of a planar surface (e.g., blank wall), the depth map degenerates to constant value, eliminating geometric guidance. However, such cases are rare in practical.  Concatenating source and target video tokens improves generation quality, but increases computational demands. Additionally, our framework cannot directly generate trajectories of 360 degree rotation, because warped depth video loses too much information at large viewpoint changes. However, an autoregressive generation manner could be leveraged to handle this situation. We leave this for future work.

\end{document}